\documentclass{article}

\usepackage{amsmath,amsfonts,bm}

\def\eqref#1{equation~\ref{#1}}

\def\1{\bm{1}}

\DeclareMathAlphabet{\mathsfit}{\encodingdefault}{\sfdefault}{m}{sl}
\SetMathAlphabet{\mathsfit}{bold}{\encodingdefault}{\sfdefault}{bx}{n}

\DeclareMathOperator*{\argmax}{arg\,max}

\usepackage{microtype}
\usepackage{graphicx}
\usepackage{booktabs} 
\usepackage{algorithm}
\usepackage{algorithmic}
\usepackage{comment}
\usepackage[normalem]{ulem}
\usepackage{caption}
\usepackage{subcaption}
\usepackage{eqparbox}
\usepackage[T1]{fontenc}

\usepackage{comment}
\renewcommand\algorithmiccomment[1]{%
  \hfill\#\ \eqparbox{COMMENT}{#1}%
}
\usepackage{xcolor}
\definecolor{deepForestGreen}{rgb}{.2,.8,.04}

\usepackage{hyperref}

\usepackage[accepted]{icml2020}

\icmltitlerunning{Enhanced POET: Open-ended Reinforcement Learning}

\begin{document}

\twocolumn[
\icmltitle{Enhanced POET: Open-Ended Reinforcement Learning through Unbounded Invention of Learning Challenges and their Solutions}



\icmlsetsymbol{equal}{*}

\begin{icmlauthorlist}
\icmlauthor{Rui Wang}{uber}
\icmlauthor{Joel Lehman}{uber}
\icmlauthor{Aditya Rawal}{uber}
\icmlauthor{Jiale Zhi}{uber}
\icmlauthor{Yulun Li}{uber}
\icmlauthor{Jeff Clune}{equal,goo}
\icmlauthor{Kenneth O. Stanley}{equal,uber}
\end{icmlauthorlist}

\icmlaffiliation{uber}{Uber AI Labs}
\icmlaffiliation{goo}{OpenAI. Work done at Uber AI Labs}

\icmlcorrespondingauthor{Rui Wang}{ruiwang@uber.com}
\icmlcorrespondingauthor{Jeff Clune}{jeffclune@OpenAI.com}
\icmlcorrespondingauthor{Kenneth O. Stanley}{kstanley@uber.com}

\icmlkeywords{Machine Learning, ICML}

\vskip 0.3in
]



\printAffiliationsAndNotice{\icmlEqualContribution} 

\begin{abstract}
Creating \emph{open-ended algorithms}, which generate their own never-ending stream of novel and appropriately challenging learning opportunities, could help to automate and accelerate progress in machine learning. 
A recent step in this direction is the Paired Open-Ended Trailblazer (POET), an algorithm that generates and solves its own challenges, and allows solutions to \emph{goal-switch} between challenges to avoid local optima. However, the original POET was unable to demonstrate its full creative potential because of limitations of the algorithm itself and because of external issues including a limited problem space and lack of a universal progress measure.
Importantly, both limitations pose impediments not only for POET, but for the pursuit of open-endedness in general.
Here we introduce and empirically validate two new innovations to the original algorithm, as well as two external innovations designed to help elucidate its full potential.  Together, these four advances enable the most open-ended algorithmic demonstration to date. 
The algorithmic innovations are (1) a domain-general measure of how meaningfully novel new challenges are,
 enabling the system to potentially create and solve \emph{interesting} challenges endlessly, and (2) an efficient heuristic for determining when agents should goal-switch from one problem to another (helping open-ended search better scale).
 Outside the algorithm itself, to enable a more definitive demonstration of open-endedness, we introduce
 (3) a novel, more flexible way to encode environmental challenges, and (4) a generic measure of the extent to which a system continues to exhibit open-ended innovation. Enhanced POET produces a diverse range of sophisticated behaviors that solve a wide range of environmental challenges, many of which cannot be solved through other means. 
 It takes a step towards producing AI-generating algorithms, which could one day bootstrap themselves from simple initial conditions to powerful cognitive machines, potentially helping with the long-term, grand ambitions of AI research.
\end{abstract}

\section{Introduction}
\label{introduction}

The progress of machine learning so far mostly relies upon a series of challenges and benchmarks that are manually conceived by the community (e.g.\ MNIST \cite{lecun1998gradient}, ImageNet \cite{deng2009imagenet}, pole balancing \cite{anderson1989learning}, and Atari \cite{bellemare2013arcade}). Once a learning algorithm converges, or solves a task, there is nothing to gain by 
running it longer in that domain. 
Sometimes, learned parameters are transferred between challenges \cite{yosinski2014transferable}. 
However, in such cases a human manually chooses which task to transfer from and to, slowing the process and limiting the opportunities to harness such transfer to cases where humans recognize its value. 
%

A fundamentally different approach is to create \emph{open-ended} algorithms \cite{standish2003open,langdon:metas05,bedau:alife08, taylor:alife16, stanley2017open,schmidhuber2013powerplay, forestier2017intrinsically} that propel \emph{themselves} forward by conceiving simultaneously \emph{both} challenges and solutions, 
thereby creating a never-ending stream of learning opportunities across expanding and sometimes circuitous webs of stepping stones. Such an algorithm also need not rely on our intuitions to determine either the right stepping stones or in what order they should be traversed to learn complex tasks, both notoriously difficult decisions \cite{stanley:book15}. Instead, it could continually invent environments that pose novel challenges of appropriate difficulty, to stimulate further capabilities without being so difficult that all gradient of improvement is lost. The environments need not arrive in a strict sequence either; they can be discovered in parallel and asynchronously, in an ever-expanding tree of diverse challenges and their solutions.

The concept of \emph{open-endedness} takes inspiration from natural evolution, which creates problems (aka challenges, niches, environments, learning opportunities, etc.), such as reaching and eating the leaves of trees for nutrition, \emph{and} their solutions, such as giraffes and caterpillars, in an ongoing process that has avoided stagnation and continued to produce novel artifacts for billions of years (and still continues). Open-endedness is also reflected in human innovation within art and science, which almost never unfold as a single linear progression of optimization aiming towards a given objective \cite{stanley:book15}. Rather, they
generate innumerable parallel and interacting branches of ideas, radiating continually in producing divergent outputs. New discoveries continue to extrapolate from their predecessors with no unified endpoint in mind. 
Open-endedness as a field of study encompasses all kinds of processes that have these properties \cite{stanley2017open,taylor:alife16}. A fascinating, challenging research question is how we can create algorithms that exhibit such open-endedness; that is, can we ignite a process that unboundedly produces and solves increasingly diverse and complex challenges (given sufficient computation)? 

The quest to achieve such open-endedness in computation has so far proven vexing \cite{taylor:alife16,dolson2018what}. First, algorithms need to maintain a delicate balance between diversity (e.g.\ pursuing different kinds of solutions simultaneously) and optimization (e.g.\ giving one arguably ``best'' solution) \cite{brant:mcc17,soros:alife14,pugh:frontiers16,lehman:gecco11, mouret:arxiv15}, as those solely focusing on optimization often lead to convergence. Second, the domain has to sustain endless opportunities to explore and learn something new. In a sense, there is a need for self-generated curricula that can continue to unfold indefinitely. (Such curriculum building has begun to emerge as its own field of study in reinforcement learning (RL), as reviewed in Section \ref{related_work}). Finally, (unbounded) innovation needs to be measured quantitatively, a problem that still lacks a satisfying solution despite some thought-provoking efforts in the past \cite{bedau:alife92}.

Recently, the \emph{Paired Open-Ended Trailblazer} (POET) algorithm \cite{wang2019paired, wang2019poetGECCO} took a step towards tackling some of these challenges (and thus towards open-ended algorithms) by simultaneously creating problems (i.e.\ learning environments) while also learning to solve them. However, while it lays a foundation for open-ended computation, the original demonstration of POET (called \emph{original POET} from here onward) still grapples with the field's longstanding challenges with balancing creativity and optimization. 
In particular, maintaining diverse skills and challenges (which supports multiple divergent streams of exploration) is critical to an effective creative process.  In POET, this diversity is enforced through a measure of environmental novelty.  Yet while this approach succeeded in its original realization, an obstacle to building on this success is that the means for measuring novelty was domain specific, which means it would in effect need to be re-designed to apply POET to any new domain.  If a genuinely domain-general approach to measuring environmental novelty could be formalized, as this paper attempts to do, it would open up POET to broad application across almost any conceivable domain with little impediment.

In addition, limitations of the original domain on which it was tested, and the lack of a general measure of open-ended progress further complicated establishing its potential. Nevertheless, the fundamental insights behind POET come tantalizingly close to pushing past the limitations of convergent optimization, which could open up a new experimental paradigm in open-ended computation.  

With these aim in mind, this paper first enhances the POET algorithm to more effectively generate and exploit diversity through two key innovations:
(1) As suggested above, instead of a hand-designed, domain-specific metric to decide the novelty of an environment, the first fully-generic method for identifying \emph{meaningfully} novel environments is formulated.  It is based on the insight that what makes an environment interesting is how agents behave in it, and novel environments are those that provide new information about how the behaviors of agents within them differ; 
(2) A more computationally efficient heuristic is formalized for determining when agents should goal-switch from one environment to another.
It also introduce two innovations that are external to the algorithm, but still crucial to demonstrating the potential of open-ended algorithms like POET: (3) a novel environmental encoding generates much more complex and diverse environments than what was used in original POET experiments,  and (4) a novel measure for quantifying open-endedness allows the claim of enhanced open-endedness to be validated objectively. As shown by experiments in this paper, the result of these four
innovations is a definitive demonstration of
open-endedness, a phenomenon rarely observed in learning algorithms.
For the field of machine learning, this kind of progress in open-ended learning is important because it offers a potential source of unbounded advancement where preexisting data is scarce or unavailable, and where the ultimate potential for discovery and achievement is unknown.

\section{Related Work}
\label{related_work}

The balance between diversity and optimization figures prominently in the field called \emph{quality diversity} (QD) \cite{pugh:frontiers16, lehman:gecco11, mouret:arxiv15}, in which the aim is to collect a diversity of high-quality solutions. Results from QD algorithms have shown that simultaneously optimizing solutions to many different problems and allowing goal-switching between tasks (i.e.\ allowing a copy of a solution being optimized for one task to switch to start being optimized to solve a different task if it looks promising for that other task) dramatically improves performance, including solving previously unsolvable problems like Montezuma's Revenge or rapid damage recovery in robots \cite{cully:nature15,nguyen2016understanding, ecoffet2019go, lehman:gecco11, mouret:arxiv15}. However, though it is closely related to open-endedness, QD does not involve the continual invention of new \emph{problems}.

Other longstanding threads of research into \emph{self-play} \cite{samuel1967some, selfplay:arxiv18, silver2018general,openai2019dota, Balduzzi2019OpenendedLI} and Generative Adversarial Networks (GANs) \cite{goodfellow2014generative} (both related to \emph{coevolution} \cite{ficici:alife98,wiegand:gecco01,popovici:hnc12}) have shown the benefit of optimizing against constantly changing, increasingly difficult challenges (e.g.\ against oneself or an opponent that also learns). 
Some recent exciting research also exists at the intersection of self-play and QD, e.g.\ AlphaStar \cite{vinyals2019grandmaster} applies extensions of population-based training \cite{jaderberg2017population} to maintain a diversity of high-quality strategies \cite{arulkumaran2019alphastar}.

Recognition of the importance of self-generated curricula is also reflected in recent advances in automatic curriculum building for RL, where the intermediate goals of curricula towards a given, final objective are automatically generated via approaches such as goal generation \cite{florensa2018automatic}, reverse curriculum generation \cite{pmlr-v78-florensa17a}, intrinsically motivated goal exploration processes (IMGEPs) \cite{forestier2017intrinsically}, teacher-student curriculum learning \cite{matiisen2017teacher}, or procedural content generation (PCG) methods (usually focused on gaming) \cite{togelius:ieeetciaig11,shaker2016procedural, justesen2017procedural}. 

Historically, a small community within the field of artificial life \cite{standish:ijcia03,langdon:metas05, lehman:alife08,graening:ppsn10,soros:alife14, soros2:alife14, brant:mcc17, stanley2017open,ray:alife91,taylor:alife16}
has studied the prospects of open-ended computation for many years. In the ongoing multi-pronged quest in pursuit of powerful AI, open-endedness is a critical prong: it could serve to generate training environments for meta-learning algorithms \cite{finn2017model,duan2016rl,wang2016lRL}, and eventually act as a stepping stone towards AI-generating algorithms (AI-GAs) \cite{clune2019ai} that could one day bootstrap themselves from simple initial conditions to powerful cognitive machines.

\section{Methods}
\label{methods}

This section first describes the original POET framework \cite{wang2019paired,wang2019poetGECCO}, and then details the two enhancements that help POET reach its potential of producing general open-ended innovation. 

\subsection{The Original POET Framework}
\label{overview_poet_framework}

\begin{figure}
    \centering
    \includegraphics[width=1.\linewidth]{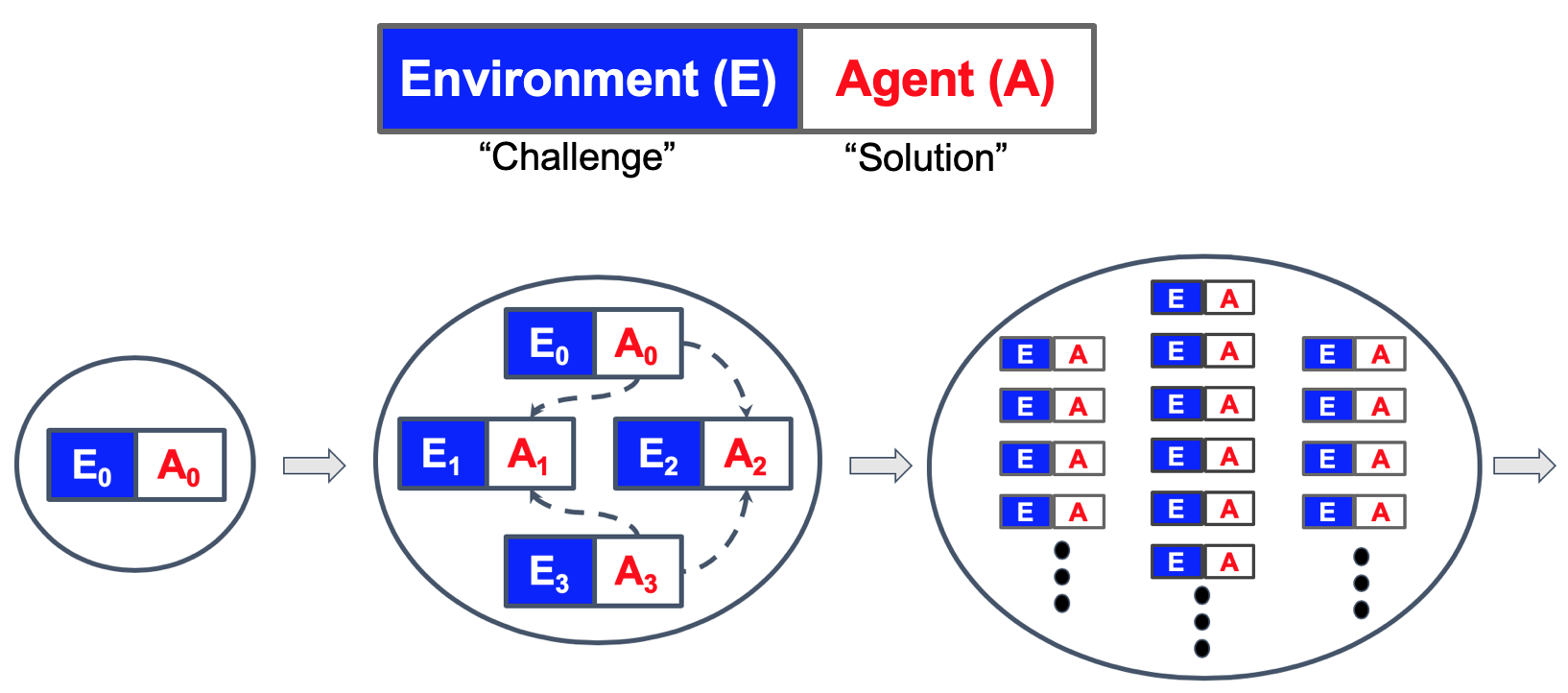}
    \caption{\textbf{POET maintains and grows a population of environment-agent pairs.} Each environment is paired with an agent being optimized to solve it. The system typically starts with a simple environment and then gradually creates and adds new environments (and their paired agents) with increasing diversity and complexity. POET harness goal-switching by periodically testing whether the current best solution to one challenge is also better than an incumbent on another challenge and, if so, replacing the incumbent with a copy of the better agent (dashed arrows). 
    }
    \label{fig:poet_system}

\end{figure}

To facilitate an open-ended process of discovery within a single run, POET grows and maintains a \emph{population} of environment-agent pairs, where each agent is optimized to solve its paired environment (Figure \ref{fig:poet_system}). POET typically starts with a trivial environment and a randomly-initialized agent, then gradually creates new environments and searches for their solutions by performing three key steps: 
\textbf{(1)} Every $M$ iterations POET generates new environments by applying mutations (i.e.\ random perturbations) to the \emph{encoding}\footnote{In original POET, the environment is encoded as a small set of hand-picked parameters. A less limited and more sophisticated encoding is introduced later in this paper.} of active environments whose paired agents have exhibited sufficient performance, signaling that perturbations of such environments are likely to be useful for encouraging learning progress. Once generated, new environments are filtered by a \emph{minimal criterion} \cite{brant:mcc17} that ensures that they are neither too hard nor too easy for existing agents in the current population, i.e.\ that they are likely to provide a promising environment for learning progress. From those that meet this minimal criterion, only the most \emph{novel} are added to the population, which pushes the environments towards capturing meaningfully \emph{diverse} learning opportunities. 
Finally, when the population size reaches a preset threshold (set in accordance with available computational resources), adding a new environment results also in moving the oldest active one from the population into an inactive \emph{archive}.  
The archived environments are then still used in the calculation of novelty for new candidate environment so that previously-existing environments are not discovered repeatedly.
\textbf{(2)} POET \emph{continually} optimizes every agent in the population within its paired environment with a variant of the evolution strategies (ES) algorithm popularized by \citet{salimans2017evolution} (in principle, any RL algorithm could be used).
\textbf{(3)} Every $N$ iterations POET tests whether a copy of any agent should be transferred from one environment to another within the population to replace the target environment's existing paired agent (i.e.\ ``goal-switching'' to solve a different task), if the transferred agent either immediately (through \emph{direct transfer}) or after one optimization step (through \emph{fine-tuning transfer}) outperforms the incumbent. In original POET \cite{wang2019poetGECCO,wang2019paired}, one of its main findings  was that such transfer is essential to finding solutions to increasingly complex and difficult environments. 

\subsection{Enhancing POET}
\label{innovations}

A central aspiration of POET is to make open-ended discovery of new problems (environments) and agents that solve them as domain-independent and efficient as possible. The first enhancement in this section makes this kind of domain independence significantly more realistic than in the original POET. 
After that, we identify and fix an inefficiency in the original POET transfer mechanism. 


When POET is applied to a particular domain, such as the obstacle courses in this paper, two important concepts are essential to the search through environments: the \emph{environmental encoding} (EE), which is a mapping from a parameter vector to an instance of an environment, creating an environmental search space, and the \emph{environment characterization} (EC), which describes key attributes of an environment that thereby facilitate calculating \emph{distances} between environments. POET harnesses this distance information to encourage the production of \emph{novel}\footnote{The novelty of an environment is calculated as the average distance to its $k$ nearest neighbors ($k = 5$ in this paper) among the active population and archive of environments \cite{lehman2011abandoning,wang2019paired,wang2019poetGECCO}.} environments. In original POET, the EE and EC are both derived from the same set of static, hand-coded features that directly tie to the domain itself (e.g.\ the roughness of the terrain and the ranges of stump heights and gap sizes). This conflation of EE and EC seems convenient, but is also a key limitation to the system's creative potential: if the EC is itself hand-coded to fit the specific domain by e.g.\ specifying fixed, preconceived properties such as a terrain's smoothness or its vertical span, then the system's output will be bound to exploration only within such prescripted possibilities. A key contribution of this paper is thus to formulate an EC that is both domain-independent and principled from the perspective of open-ended innovation.

\begin{figure}[!ht]
    \centering
    \begin{subfigure}[b]{1.\linewidth}
        \centering
        \includegraphics[width=.75\linewidth]{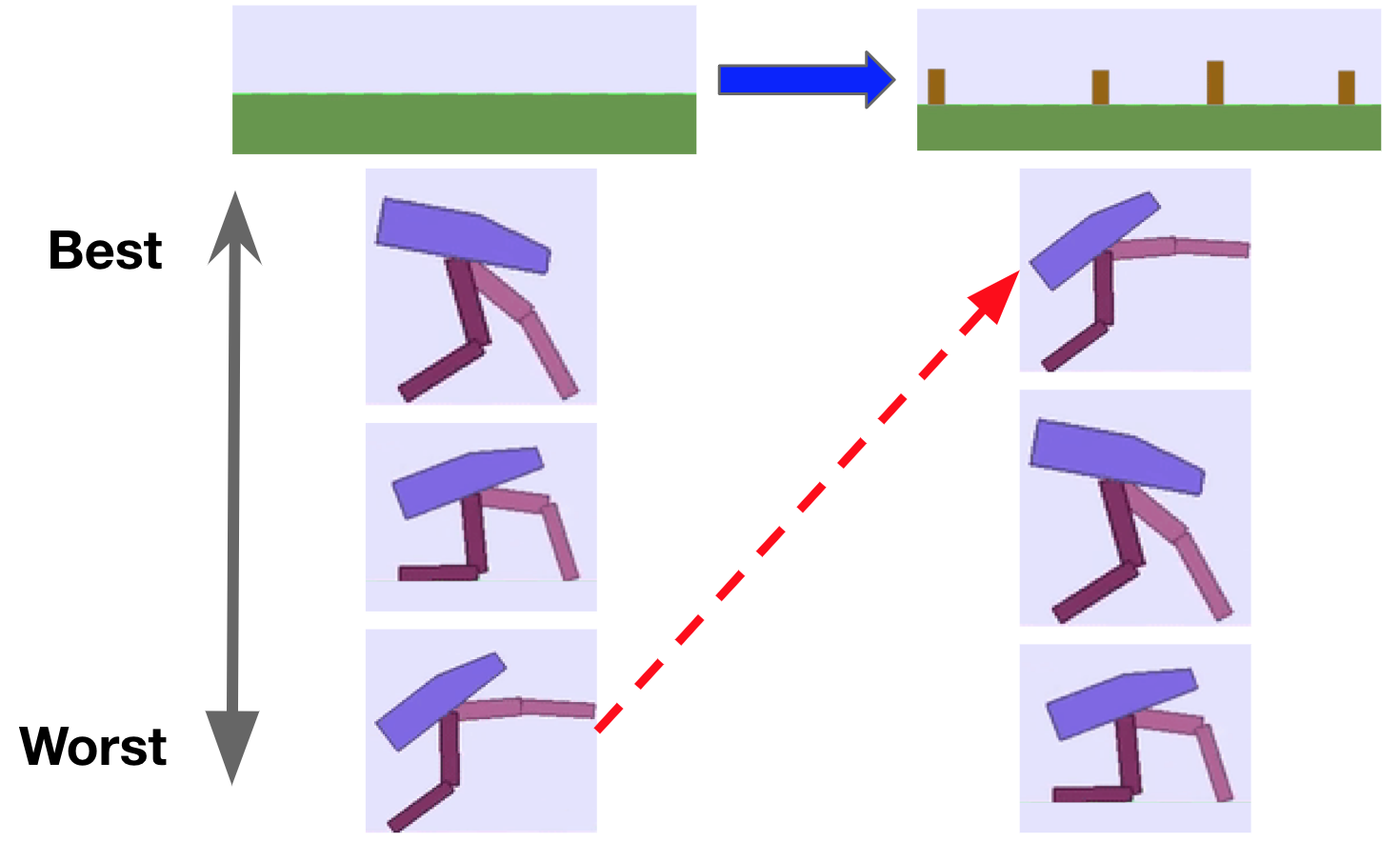}
        \caption{The emergence of stumps induces different orderings of agents.
        }
        \label{fig:why_t_vec}
    \end{subfigure}
    ~
    \begin{subfigure}[b]{1.\linewidth}
        \includegraphics[width=\linewidth]{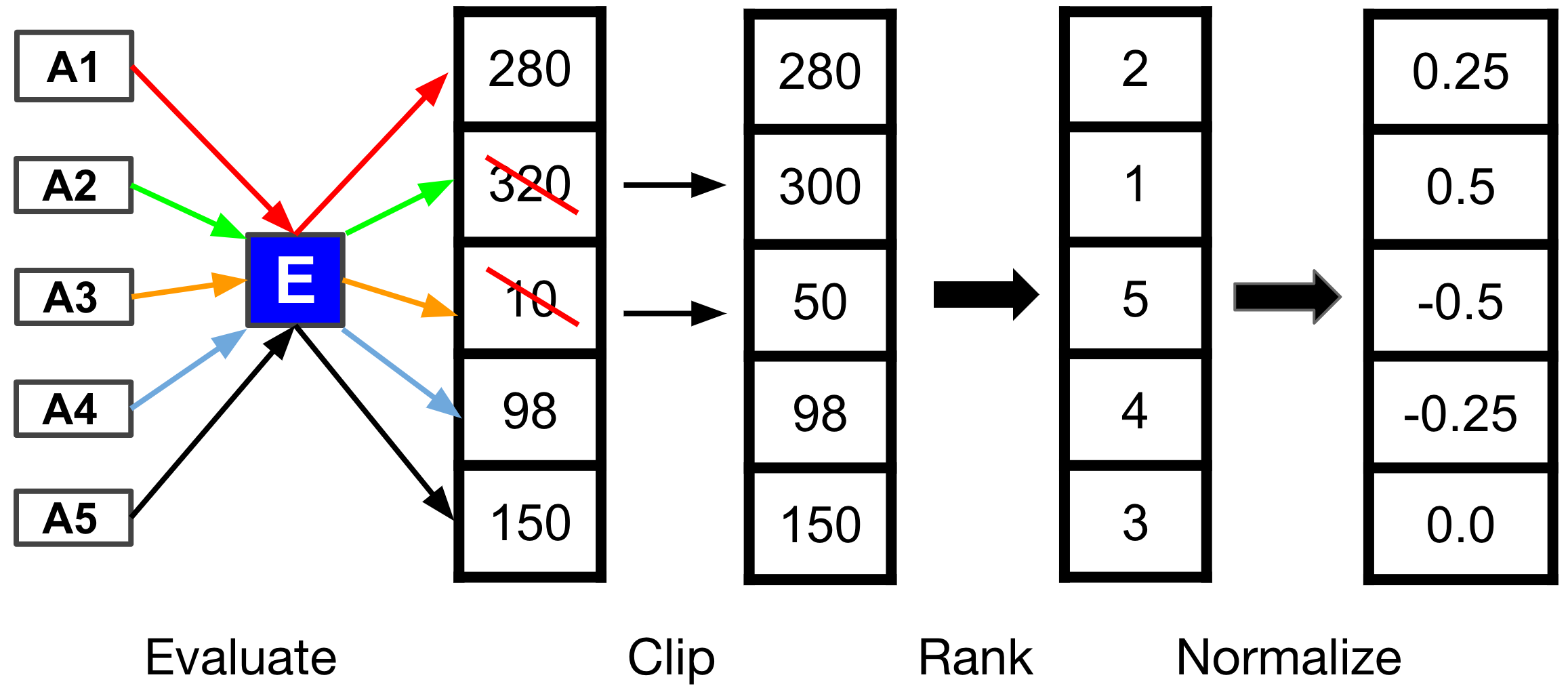}
        \caption{Steps to calculate the PATA-EC for an environment (E).}
        \label{fig:how_t_vec}
    \end{subfigure}    
    \caption{\textbf{PATA-EC, a domain-general distance metric for measuring meaningfully different environments.} (a) An agent that walks with one leg raised is not energy-efficient on flat ground and thus ranks last, but that gaits enables it to step over high stumps and thus ranks highest in a more stumpy environment. (b) The calculation of the PATA-EC for environment E based on the rank of performance of five agents (A1--A5).
}
    \label{fig:why_how_t_vec}
\end{figure}

Our proposed domain-general EC, the \emph{Performance of All Transferred Agents EC} (PATA-EC), is grounded by how all agents (in the population and archive) perform in that environment. The key insight motivating the PATA-EC is that a novel and useful challenge should make novel \emph{distinctions} among agents in the system \cite{Jong2004IdealEF}: if a newly-generated environment induces a significantly distinct \emph{ordering} on how agents perform within it (relative to other environments), it likely poses a qualitatively new kind of challenge. For example, as illustrated in Figure \ref{fig:why_t_vec}, the emergence of a new landscape with stumps may induce a new ordering on agents, as agents with different walking gaits may differ on their ability to step over protruding obstacles. An important aspect of this insight, based on how environments order agents, is that it does not rely upon any domain-specific information at all.

Figure \ref{fig:how_t_vec} illustrates the steps to calculate the PATA-EC for any given environment: (1) \textbf{Evaluate}: Each environment evaluates all active and archived 
agents and stores their raw scores in a vector. Note that the required computation already occurs incidentally in the course of POET for active environments as a result of the  transfer mechanism (which tries agents in their non-native environments). (2) \textbf{Clip}: Each score in the vector is clipped between a lower bound and an upper bound. 
The intuition is that both extreme scenarios are irrelevant for learning progress: a score that is too low indicates outright failure of an agent, while a score that is too high hints that an agent is already competent. (3) \textbf{Rank-normalize}: The PATA-EC is assembled by replacing the scores with rankings, and then normalizing each rank to the range of $[-0.5, 0.5]$ (similar techniques have been adopted in the ES literature, e.g.\ \citet{wierstra2014natural,salimans2017evolution}). Performing this normalization allows direct use of the Euclidean distance metric to measure the distance between PATA-ECs, which worked empirically better than rank-correlation-based distance metrics directly defined on vectors of rankings in preliminary tests. The implication of the PATA-EC is significant: We can now measure and reward environmental novelty completely independently of any domain-specific information, opening up POET to almost any conceivable domain.


Finally, POET's \emph{transfer mechanism} enables innovations from solutions for one environment to aid progress in other environments by periodically attempting to replace an incumbent agent (for a target environment) with another agent in the population that performs better in that environment. While critical for the overall performance, the transfer mechanism in original POET also creates two problems: it is (1) computationally expensive because it involves an optimization step to compute the fine-tuning transfer score for each transfer evaluation, and (2) prone to ``false positives'' due to stochasticity in RL optimization and a low bar for replacing more proven incumbents. To effectively remedy both pitfalls, we introduce a more stringent threshold (i.e.\ the maximum of the 5 most recent scores of the incumbent) that \emph{both} direct and fine-tuning transfer scores (instead of either one, as in original POET) of a candidate agent must exceed to qualify as an incoming transfer (Algorithm \ref{alg:eff_transfer} in Appendix \ref{appendix:improved_transfer_algo}). This simplification not only smooths out noise from the stochasticity of ES optimization, but also saves computation because the fine-tuning step is only performed if the direct transfer test is passed. 

Now with the enhanced algorithm at hand, uncovering its full potential will require two additional innovations
extrinsic to the algorithm itself.

\section{More Expressive Environment Encoding}
\label{subsec:ee}

Even with the right algorithm, innovation will eventually grind to a halt if the domain itself is limited. The challenge lies in how to formalize an encoding that can sustain an environmental space with possibilities beyond the imagination of its designer. 
In original POET, the 2-D bipedal walking environments are encoded by a fixed, small set of hand-picked parameters (e.g.\ ranges of stump height and gap width, surface roughness, etc.) that can only support a finite number of obstacle types with predefined regular shapes and limited variations (e.g. Figure \ref{fig:poet1_env}). While this encoding expresses sufficient possibilities for POET to demonstrate an initial period of innovation, such innovation by necessity will eventually peter out as possible novel environments to explore gradually run out. 
To overcome this limitation,
a desired encoding should be highly \emph{expressive}, i.e.\ able to express environmental details with a high degree of granularity and precision to capture ever-more-intricate detail. 

\begin{figure}
\centering
\includegraphics[width=1.\linewidth]{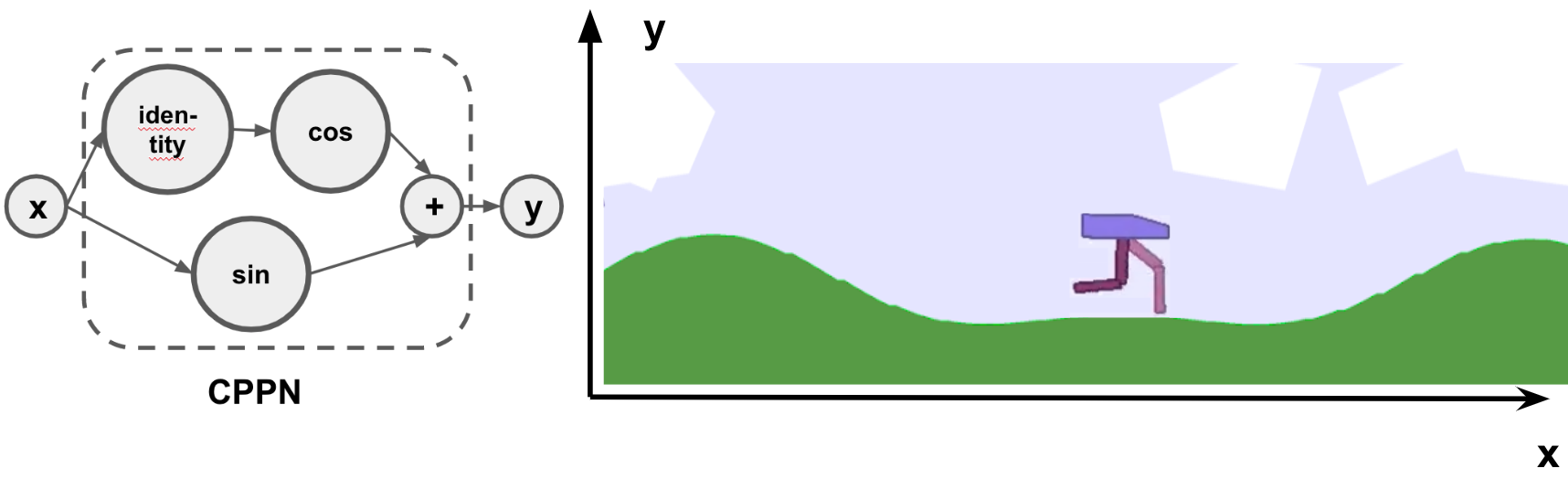}
\caption{\textbf{A sample CPPN (left) and its generated landscape (right).} The CPPN produces $y$ coordinates, given each $x$ coordinate, which are then rendered into a bipedal walker environment for the Bipedal Walker environment in OpenAI Gym \cite{brockman2016openai}. An agent, shown in the right figure, is controlled by a different agent neural network to navigate through the generated landscape and is rewarded for quickly moving from left to right.
}
\label{fig:cppn}
\end{figure}

A class of neural networks known as compositional pattern-producing networks (CPPNs) \cite{stanley2007compositional} are a candidate for a general encoding mechanism that respects this requirement. 
CPPNs take as input geometric coordinates (e.g.\ $x$ and $y$), and when queried across such coordinates produce a geometric pattern (e.g. 2-D images). 
Figure \ref{fig:cppn} illustrates how to generate the landscape of a 2-D bipedal walking environment from a single-input, single-output CPPN, which is queried across the space of $x$ coordinates that compose the landscape. Its output is interpreted as the height of the landscape at that point. (A selection of more complex CPPN landscapes from POET runs are later shown in Figures \ref{fig:poet2_env} and \ref{fig:landscape_12}.) As an encoding mechanism, CPPNs offer desirable properties for open-endedness: (1) They are typically initialized with simple topologies (e.g.\ no hidden nodes), and are trained with NEAT \cite{stanley:ec02}, a neuroevolution \cite{stanley2019nature} algorithm that learns both the topology and the weights of CPPNs (details in Appendix \ref{appendix:cppn-details}). As a result, simple (e.g.\ flat or sloped) landscapes are often produced in the beginning of a POET run, while more complex (and often more challenging) landscapes gradually emerge as NEAT’s topology-altering mutations (e.g.\ adding a node or a connection) gradually elaborate the neural architecture of the CPPNs. (2) Because CPPNs can evolve arbitrarily complex architectures, in this domain they can in theory express any possible landscape at any conceivable resolution or size.  

It is important to note that it is \emph{because} of the generic PATA-EC that we are now able to measure diversity with respect to CPPN-generated levels, for which otherwise there is no obvious principled approach.
The idea is that this novel, more expressive way to encode and create environments, coupled with the generic EC described in the previous section, significantly increases the potential for POET to exhibit open-ended innovation compared to the simple, fixed encoding from the original POET experiments.

\section{The ANNECS Measure of Progress}
\label{novel_measure}

Measuring progress in open-ended systems has long presented a challenge to pursuing open-endedness: As there is no a priori expected outcome against which progress can be measured, how can we tell whether a system continues to generate \emph{interesting} new things? 
The new idea here is that measuring progress can be based on 
the idea that
if an existing set of agents are able to solve all of the new challenges generated by a system in the future, then the system has not generated any meaningfully new challenges. 
The system also should not generate problems with no hope of being solved.
Therefore, we propose to track the \emph{accumulated number of novel environments created and solved} (ANNECS) across the duration of a run of an open-ended system. Specifically, to be counted in ANNECS, an environment created at a particular iteration (1) must pass the minimal criterion (i.e.\ that it is neither too hard nor too easy) measured against all the agents (in
cluding ones currently in the active population and in the archive) generated over the entire current run so far, 
and, (2) must be eventually solved by the system (which means that the system does not receive credit for producing unsolvable challenges). 
This proposed metric ties directly to the overall effectiveness of an open-ended process: As the run proceeds, the ANNECS metric consistently going up indicates that the underlying algorithm is constantly creating meaningfully new environments. 


\section{Experiments and Results}
\label{experiments}

With an enhanced algorithm, a more open-ended environment encoding, and a new means for measuring open-ended innovation over time, the question now is whether a definitive improvement in open-ended computation can be demonstrated.  The aim in this section is to attack this question from several angles, both to show why open-endedness remains unique in its potential among all the methods in machine learning, and also how the enhancements to POET genuinely improve its tendency towards continual innovation.  

For this purpose, the experimental approach is to empirically evaluate Enhanced POET in a
domain adapted from the 2-D bipedal walking environment used in the original POET, which itself was based on the ``Bipedal Walker Hardcore'' environment of OpenAI Gym \cite{brockman2016openai}.
An instance of this experimental domain consists of a bipedal walking agent and an obstacle course that the agent attempts to navigate from left to right (Figure \ref{fig:cppn}). The agent in this work has the same configuration as in the original POET 
(also described in Appendix \ref{appenidx:additional_domain_details}), while the obstacle courses now can be encoded and generated by the CPPN-based EE (Section \ref{subsec:ee}). The experiments are organized to demonstrate the values of the four main contributions: we first evaluate the performance of the new EC and improved transfer strategy, respectively, and then test the overall performance of the Enhanced POET with the CPPN-based EE. Lastly, the new ANNECS metric is put to the test, measuring progress in Enhanced POET and contrasting it with the original POET (Section \ref{novel_measure}). Unless noted otherwise, a POET run takes 60,000 POET iterations with a population size of 40 active environments. Because POET consists of many independent operations, such as agents optimizing within their paired environments, as well as transfer attempts, it is feasible and favorable to distribute the computations over many processors. Our software implementation, which has been released as open source code at \url{https://www.github.com/uber-research/poet}, completes a 60,000-iteration
POET run in about 12 days with 750 CPU cores. The implementation is based on Fiber \cite{zhi2020fiber}, a recently released distributed computing library in Python that enables seamless parallelization over any number of cores. Further details about the domain and experiment setup are in Appendix \ref{appedix:exp_setup}.




For the purpose of analyzing the quality of results, this work adopts the definitions of challenge levels for the simple, hand-designed EE (i.e.\ a vector of values that consists of the surface roughness, the range of stump height, and the range of gap width) from the original POET paper \cite{wang2019paired, wang2019poetGECCO}, where environments are classified as either \emph{challenging}, \emph{very challenging}, or \emph{extremely challenging},  based on how many conditions they satisfy out of the three listed in Table \ref{challenge-table}.
Satisfying one of the three conditions makes an environment \emph{challenging}; satisfying two of the three conditions makes an environment \emph{very challenging}; and an \emph{extremely challenging} environment satisfies all three conditions.
Shown in Table \ref{challenge-table}, each of these conditions is much more demanding than the corresponding values used in the original Bipedal Walker Hardcore environment \cite{OpenAI_bipedal} in OpenAI Gym \cite{brockman2016openai}.

\begin{table*}[!ht]
\begin{center}
\begin{small}
\begin{sc}
\begin{tabular}{lccccc}
\hline
 & Maximum &  Maximum & \\
 & Stump Height & Gap Width & Roughness\\
\hline
POET & $\geq2.4$ & $\geq6.0$ &  $\geq4.5$\\
Original Bipedal Walker Hardcore environment & $2.0$ & $3.0$ & $1.0$\\
\hline
\end{tabular}
\end{sc}
\end{small}
\end{center}
\caption{\textbf{The challenge level of an environment is based on how many conditions it satisfies out of the three listed here.} Satisfying one, two or all three conditions makes an environment challenging, very challenging, or extremely challenging, respectively \cite{wang2019paired, wang2019poetGECCO}.  The values used in the experiments of POET to determine the challenge level of an environment are 1.2, 2.0, and 4.5 times the corresponding values used in the original Bipedal Walker Hardcore environment in OpenAI Gym \cite{OpenAI_bipedal}. 
\label{challenge-table}
}
\vskip -0.1in
\end{table*}

The first experiment tests whether the proposed PATA-EC indeed encourages creating and solving a diverse set of challenges. When applied to the domain in the original POET (still with the original, hand-crafted EE), we find that PATA-EC can produce the same diversity and challenge levels of environments as the original hand-designed EC, although it requires $82.4 \pm 7.31 \%$ more computation, measured in ES steps (details in Appendix \ref{appendix:figure_4}). 
Because the original EC was hand-designed for this specific domain and encoding, its performance is the best we could reasonably expect from any EC in this domain. It is therefore promising that adopting PATA-EC with full generality only carries the price of less than a factor of two slowdown, but frees us to incorporate much richer EEs (such as CPPNs) into POET, thus powering the exploration of novel, unanticipated terrains.

The second experiment evaluates the proposed improved transfer strategy. With the same setup as in original POET \cite{wang2019paired, wang2019poetGECCO} but with the improved transfer strategy,  POET can create  (and solve) the same diversity and challenge levels of environments with only $79.7\pm1.67\%$ of the computation (measured in number of ES steps) (details in Appendix \ref{appendix:figure_5}). This result suggests that the improved transfer strategy successfully reduces the cost of goal-switching in original POET without sacrificing its benefits with respect to solution discovery. 

\begin{figure*}[htp]
  \centering
  \includegraphics[width=1.\linewidth]{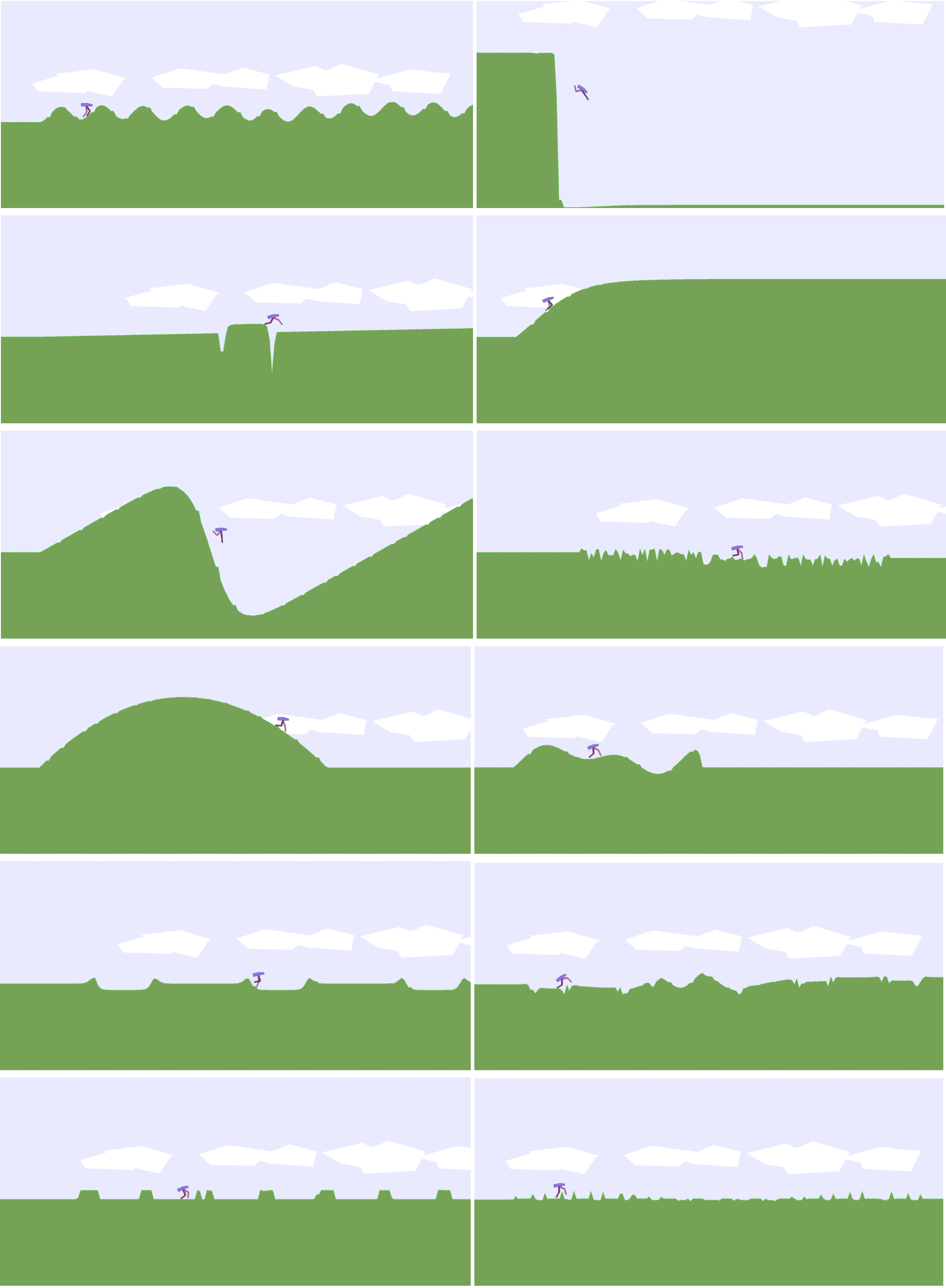}
\caption{\textbf{Sample environments created and solved in a single run by Enhanced POET with the CPPN-based EE.} These environments exhibit a wide diversity of macro and micro environmental features.
}
  \label{fig:landscape_12}
\end{figure*}

The next set of experiments leverage all four contributions of this work. We first show that Enhanced POET with the new CPPN-based EE is able to create and solve a large diversity of environments within a single run. These are qualitatively different than those produced by the original POET with the simple, hand-designed EE that supports only a few types of simple obstacles (e.g., stumps and gaps) (Figure \ref{fig:poet1_env}; more in Figure \ref{fig:original_landscape_12} in Appendix \ref{appendix:app_landscape_samples}). The CPPN-encoded environments produced by the Enhanced POET exhibit a wide variety of obstacles vastly different in their overall shapes, heights, fine details, and subtle variations (Figure \ref{fig:poet2_env} and Figure \ref{fig:landscape_12} with videos of agents at \url{https://www.youtube.com/playlist?list=PLxWSC7x4MS2feqPL7MojvfwgaQzOwk_b4}; details in Appendix \ref{appendix:app_landscape_samples}). Such diversity is also reflected in phylogenetic trees (aka family trees) of the environments it has created, which exhibit a clear signature of open-ended algorithms: multiple, deep, hierarchically nested branches, resembling those from natural phylogenies (Figure \ref{fig:phy_tree}; details in Appendix \ref{appendix:phylogenetic_tree}). It is also interesting to see that POET agents tend to be specialized to particular environments that pose very different challenges, as illustrated in the matrix formed by the vectors of scores of agents across all the first 80 environments created and solved in a POET run (Figure \ref{fig:heatmap} in Appendix \ref{appendix:pata_matrix}). 

\begin{figure}[!ht]
  \centering
  \begin{subfigure}[b]{\columnwidth}
    \includegraphics[width=\linewidth]{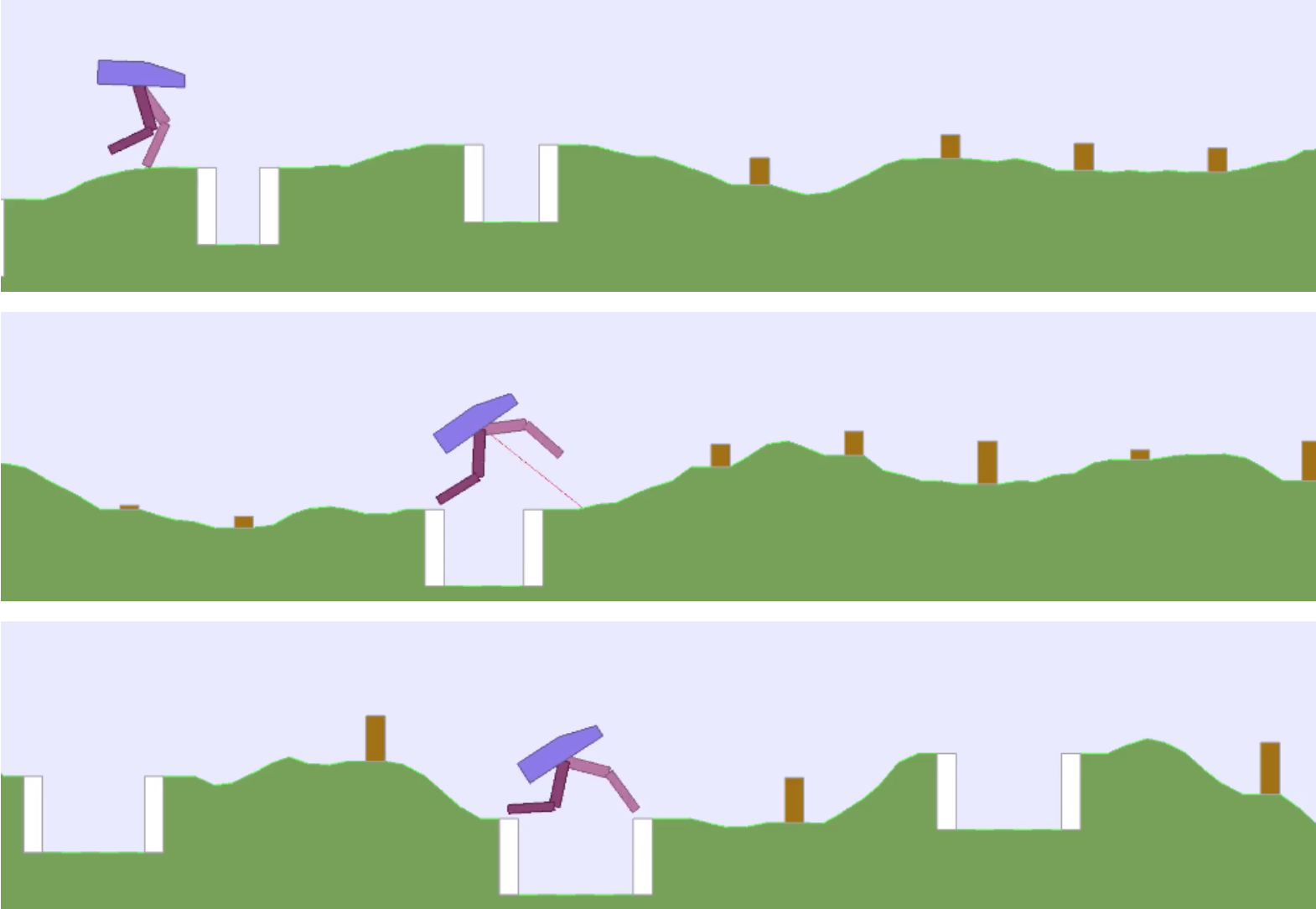}
    \caption{Sample environments from a single run of original POET.}
    \label{fig:poet1_env}
  \end{subfigure}  
  ~
  \begin{subfigure}[b]{\columnwidth}
    \includegraphics[width=\linewidth]{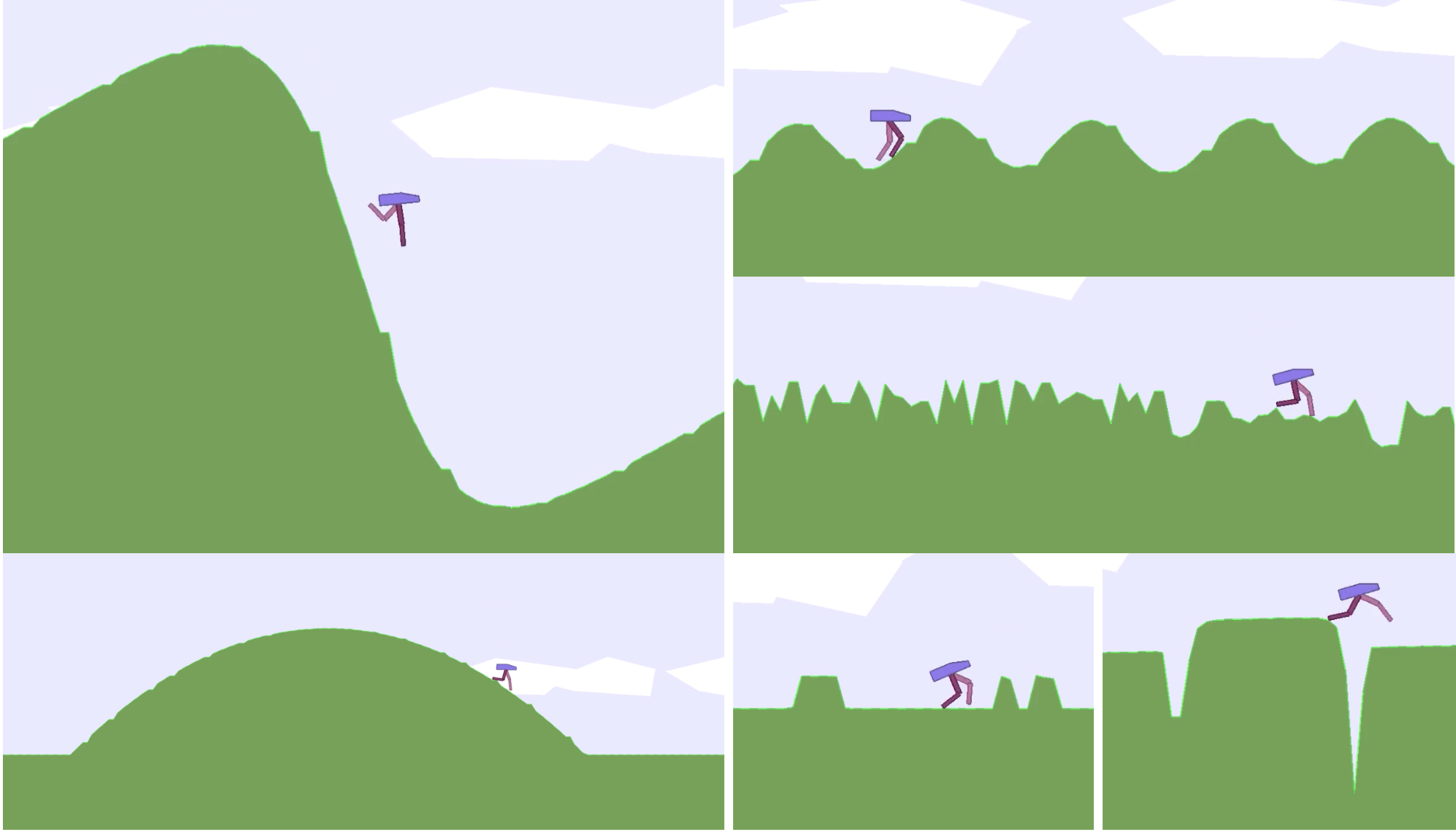}
    \caption{Sample environments from a single run of Enhanced POET.}
    \label{fig:poet2_env}
  \end{subfigure}
\caption{\textbf{With the CPPN-based EE and other innovations, Enhanced POET is able to generate (and solve) a wide diversity of environments within a single run.} In contrast, the original POET can only generate environments with limited types of regularly-shaped obstacles (e.g.\ stumps and gaps). 
}
  \label{fig:taste_landscape}
\end{figure}

\begin{figure*}[!ht]
  \centering
  \includegraphics[width=1.\linewidth]{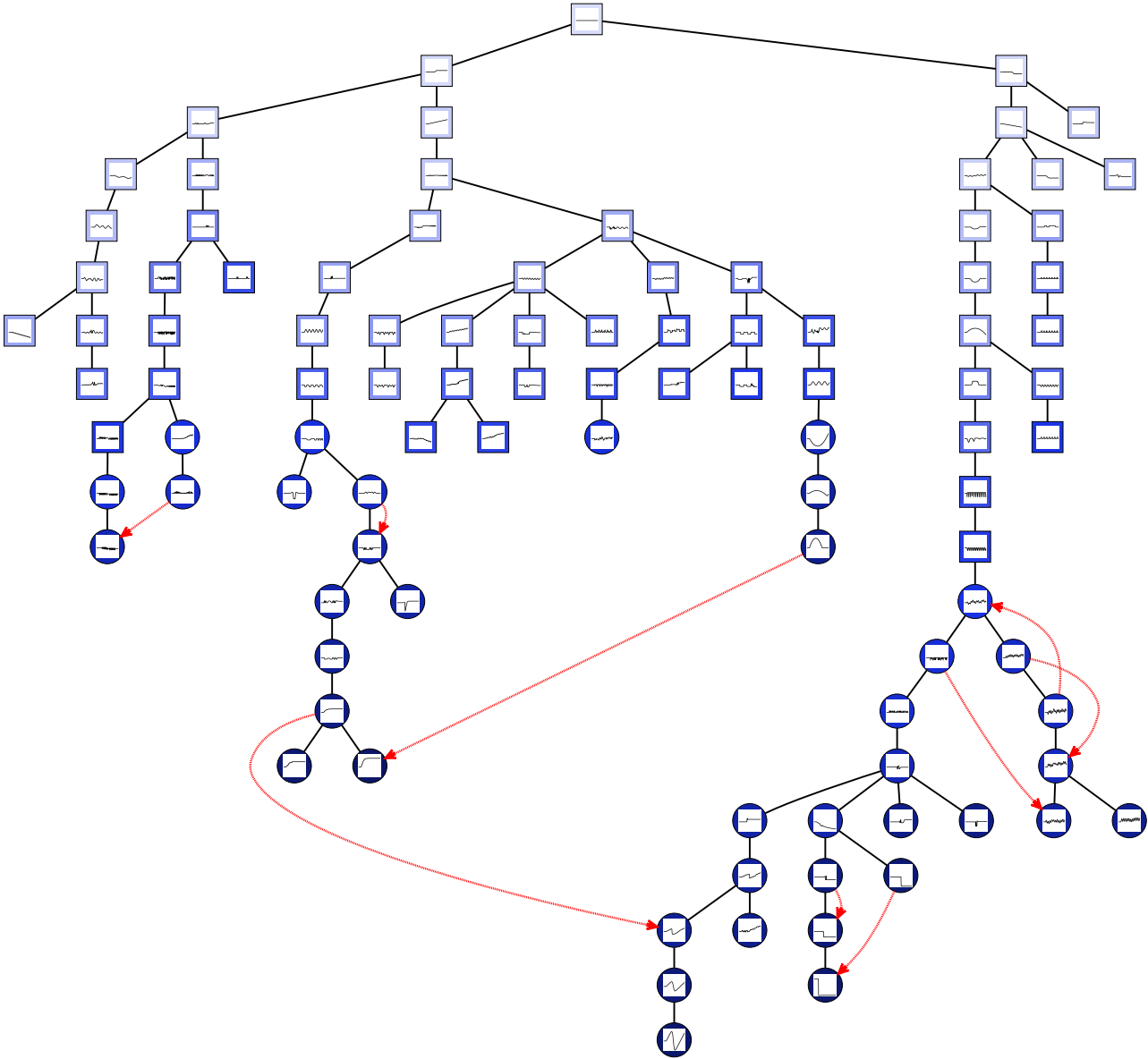}
\caption{\textbf{Phylogenetic tree of the first 100 environments of a POET run.} Each node contains a landscape picture (zooming in the digital version enables seeing more detail) depicting a unique environment, with outgoing edges on its bottom connecting to its children. The circular or square shape of a node indicates that the environment is in the active population or the archive, respectively, while the color of the border of each node suggests its time of creation: darker color means being created later in the process. The red arrows label successful transfers during a single transfer iteration, specifically between the addition of the 100th and 101st environment.
}
  \label{fig:phy_tree}
\end{figure*}


We next test an intriguing hypothesis that was also investigated for the original POET: Is it the case that some of the environments Enhanced POET generates are challenging enough that the curriculum it self-generates is \emph{necessary} to solve them? This question is interesting because it implies that modern learning algorithms on their own may be able to achieve much more than is currently known if only they were embedded into an open-ended process like POET. Our approach is to sample environments created and solved throughout Enhanced POET runs and attempt to solve them with control algorithms. Specifically, we sort all the environments generated and eventually solved in a POET run in the order of when they are solved, and select one from the first $10\%$, one from the middle ($45\%-55\%$), and one from the last $10\%$ of the run, in each case choosing the environment with the lowest initial score from that part of the run (indicating difficulty). These are referred to as \emph{early stage}, \emph{middle stage}, and \emph{late stage} environments, respectively. The process was repeated for 5 independent POET runs (each with a different random seed) to obtain in total 15 environment targets. For each target environment, two different types of controls are attempted: One is direct optimizations by ES (with the same hyperparameters as in POET) and by the Proximal Policy Optimization (PPO) algorithm \cite{schulman2017proximal} (hyperparameters in Appendix \ref{appendix:PPO_setup}), respectively. The other, stronger control is to manually create an explicit curriculum by introducing a scaling factor that multiplies the height of the ground at each position from left to right, and increase the scaling factor from 0.0 to 1.0 at a step size of 0.02. Doing so smoothly morphs the perfectly flat environment to a given target environment, yielding a natural curriculum (referred to later as the \emph{ground-interpolation curriculum}) that is analogous to the direct-path curriculum in the original POET paper.

\begin{figure}[!ht]
  \centering
    \begin{subfigure}[b]{\linewidth}
        \centering
        \includegraphics[width=\linewidth]{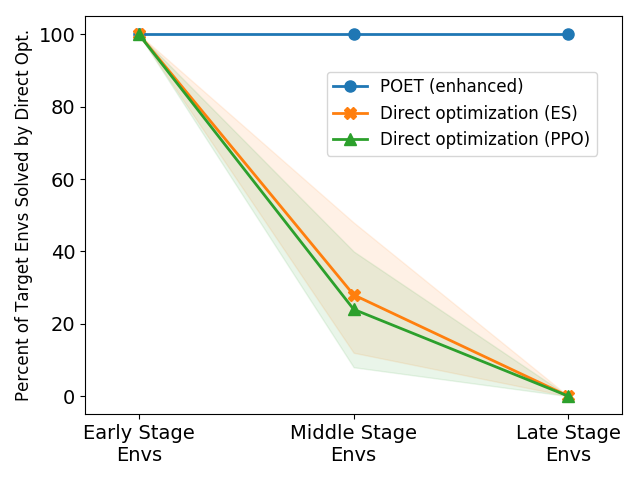}
        \caption{Percentage of target environments solved by direct optimization. Symbols: mean. Shaded regions: $95\%$ bootstrapped confidence interval.}
        \label{fig:zeroone_direct_control}
    \end{subfigure}
    \begin{subfigure}[b]{\linewidth}
        \centering
        \includegraphics[width=\linewidth]{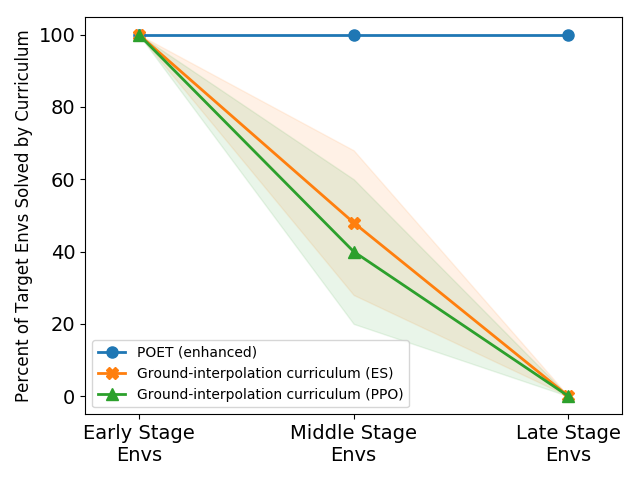}
        \caption{Percentage of target environments solved by the ground-interpolation curriculum. Symbols: mean. Shaded regions: 95\% bootstrapped confidence intervals.}
        \label{fig:zeroone_curr_control}
    \end{subfigure}
\caption{
\textbf{POET generates and solves challenges that neither direct optimization nor the ground-interpolation curriculum can solve.} Target environments were selected from different stages of POET runs (see text) and are all solved by POET (blue). For both direct optimization (a) and ground interpolation (b), controls with ES (orange) and PPO (green) respectively, can only solve those selected at the earlier stages of a POET run (which are therefore often less challenging), but could not solve more challenging target environments selected at later stages of POET runs. 
}
  \label{fig:percentage_of_solved}
\end{figure}

When given an equivalent computational budget to what POET spent to solve each target (details in Appendix \ref{appendix:equal_budget}), the two types of controls can solve target environments selected at the earlier stages of POET runs (when the produced environments are often less challenging), but both significantly underperform POET in solving middle and late stage target environments ($p < 0.01$; Wilcoxon signed rank test). Figure \ref{fig:percentage_of_solved} illustrates percentages of target environments at different stages solved by the two types of controls, respectively,  with more results given in Appendix \ref{appendix:additional_control}.
The result that neither direct optimization nor manually created curricula come close to producing the level of success in challenging environments that POET achieves via its self-generated implicit curriculum confirms the result of the original POET paper in a new setting (and now also with PPO). Interestingly, much research effort is spent attempting to design or learn single-path curricula to help an agent learn a complex task \cite{gomez1997incremental,bengio2009curriculum,karpathy2012curriculum,heess2017emergence, justesen2017procedural}.
Here, we see (again) that such efforts often do not work. POET, however, is not trying to create any specific curriculum, but ends up producing many effective curricula (within one run) to solve many different challenging tasks. It does so because it collects an ever-expanding set of stepping stones (in the form of challenges and solutions) and allows goal-switching between them, which captures serendipitous discoveries as they occur \cite{stanley:book15,nguyen2016understanding,lehman:gecco11}. 

\begin{figure}
  \centering
  \includegraphics[width=\linewidth]{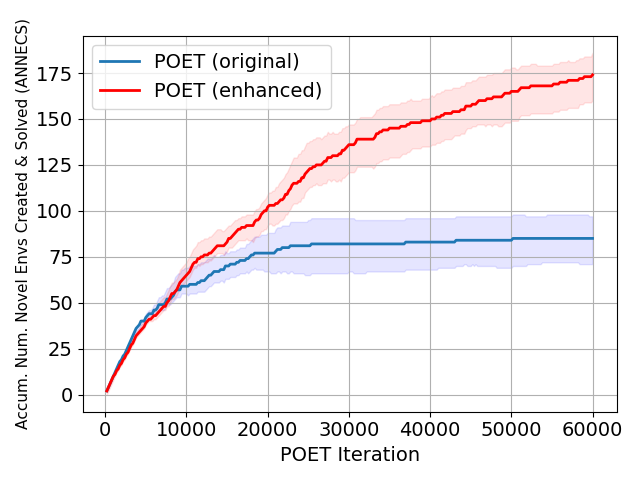}
\caption{\textbf{A comparison of the ANNECS metric across iterations between Enhanced POET and original POET.} Solid lines denotes the median across 5 runs and shading denotes the 95\% bootstrapped confidence intervals. Original POET runs gradually lose steam and plateau after around 20,000 iterations; in sharp contrast, Enhanced POET runs maintain momentum with the ANNECS consistently going up without much signs of slowing down.}
  \label{fig:capacity_curve}
\end{figure}

Finally, Figure \ref{fig:capacity_curve} compares the ANNECS metric of progress proposed in Section \ref {novel_measure} between the Enhanced POET with the new EE, and the original POET with the original (fixed) EE, a comparison that demonstrates the overall impact of both algorithmic innovations and the enhanced encoding. The original POET initially exhibits comparable performance to Enhanced POET, but eventually loses its ability to innovate, as shown by its ANNECS curve plateauing after 20,000 iterations. Such stagnation occurs because the EE for original POET can only sustain a finite number of obstacle types with predefined regular shapes and limited variations, so it gradually runs out of possible novel environments to explore. In intriguing contrast, innovation within Enhanced POET continues almost linearly throughout the experiments, though at a slightly slower speed beyond 30,000 iterations. This slight slowdown reflects that as environments generally become more challenging, it requires more optimization steps for environment-agent pairs to reach the score threshold for generating new environments (data not shown). Despite that, new environments that can pass the MC continue to be consistently discovered, significantly exceeding the duration of time that the original POET can continually innovate. The result validates the ANNECS approach by aligning with our expectation that a limited encoding cannot support long-term innovation, while the longer chain of innovation of Enhanced POET is achieved because the CPPN-encoded environmental space offers significantly more potential for meaningful diversity. Furthermore, the domain-general PATA-EC and improved transfer strategy make it possible and efficient to explore, create and solve novel environments in such a space. 

\section{Discussion, Conclusion, and Future Work}
\label{conclusion}
The reason open-endedness is so compelling and so important to the future of machine learning is suggested by an intriguing result in this paper: the very same optimization algorithm, i.e.\ ES (and PPO too),  that \emph{cannot} solve any late-stage environment from POET runs, actually \emph{can} solve them, but only if it is embedded within an open-ended \emph{algorithmic context} (in this case, POET). This result, perhaps counterintuitive at first glance, 
rests on the insight that we cannot know in advance the stepping stones that must be crossed to reach a far-off achievement. 
Science's history  repeatedly confirms this kind of lesson: Microwaves were invented not by food-heating researchers but by those studying radar; and computers were not invented by optimizing the abacus to increase computations per second, but because scientists invented vacuum tubes and electricity for entirely unrelated purposes \cite{stanley:book15}. Open-ended processes \emph{embrace} this lesson by collecting stepping stones from innumerable \emph{divergent branching} paths through the search space, many climbing towards higher complexity and challenge simultaneously, towards otherwise inconceivable future achievements.

This divergent branching in Enhanced POET is enabled by the newly-introduced PATA-EC, which resonates with ``behavior characterizations'' aimed at encouraging divergence and exploration, e.g.\ in novelty search \cite{lehman2011abandoning}, QD \cite{pugh:frontiers16} and similarly-oriented work in intrinsic motivation in RL \cite{oudeyer2009intrinsic,schmidhuber2010formal,bellemare2016unifying}. 
However, unlike previous such characterizations that struggle with the problem of domain generality, PATA-EC is an entirely general characterization, an interesting side-effect of coevolving both environments and agents together. It is precisely because we now have a palette of environments from which to sample that we can begin to construct a profile of behavior (for both environments and agents!) based on their interactions without knowing anything about the inner workings of those environments or agents.  Thus, in a sense, the push for divergence in learning (and ultimately, towards open-endedness) becomes fundamentally more tractable when environments are not predefined but instead being learned as agents are being optimized.

Yet despite all these new possibilities, a shadow of doubt lingers at the heart of open-endedness: Because we have no way to know what it may find or what the future of any given run may bring, a skeptic regarding this \emph{uncertainty} might interpret a system like POET as a kind of meandering walk through problem space dangerously close to randomness. Yet the qualitative results conflict with such a pessimistic interpretation -- indeed, these agents have gained the ability to traverse extreme irregularity underfoot (reminiscent of dried lava flows near volcanoes), to walk swiftly in efficient alternating bipedal fashion on flat ground, and even to brace remarkably for landing after a fall from great heights. Not only that, but there seems to be no other viable method to learn such skills from scratch. They are not arbitrary skills, but genuinely meaningful, sometimes beyond what we might even expect possible. If we embrace such uncertainty in algorithms that will take us to amazing places, but will not tell us our destinations ahead of time, we might harness the power and reap the rewards of powerful, open-ended search processes.  

Finally, how long might such algorithms endure? Is even a tiny sliver of the multi-billion-year saga of unfolding life on Earth even conceivable in computation? Looking at Figure \ref{fig:capacity_curve}, though clearly more enduring than its predecessor,
the curve for Enhanced POET appears to surrender to slightly more modest growth after 30,000 iterations. Is it petering out, though just more slowly?  In fact (as noted also in Section \ref{experiments}), the slower growth in ANNECS is the result of the environments becoming increasingly difficult, and thus each challenge requiring more time to optimize before more can be generated. That is different from a case where there is simply no more room for discovery in the space of the domain itself.  Yet even so, it seems inevitable in such a relatively simple world that the time will come where nothing more can be invented that is physically possible to traverse for our agent.  The ANNECS curve might be expected to flatline then.

However, that fate is not inevitable by virtue of the algorithm itself.  Rather, it seems an artifact of the domain, even when enhanced with CPPNs -- somehow, the idea of obstacle courses will succumb to its own finitude in a way that life on Earth has not.  There is a sense though in which this realization is exciting -- Enhanced POET itself seems prepared to push onward as long as there is ground left to discover.  The \emph{algorithm} is arguably unbounded.  If we can conceive a domain without bounds, or at least with bounds beyond our conception, we may now have the possibility to see something far beyond our imagination borne out of computation alone.  That is the exciting promise of open-endedness.



\section*{Acknowledgements}
We thank all of the members of Uber AI Labs, in particular Joost Huizinga, Thomas Miconi, Paul Szerlip, Lawrence Murray, and Jane Hung for helpful discussions. We are grateful to Leon Rosenshein, Joel Snow, Thaxton Beesley, the Colorado Data Center team and the entire Opus Team at Uber for providing our computing platform and for technical support.

\bibliography{example_paper,ucf,iclr2020_conference}

\begin{thebibliography}{74}
\providecommand{\natexlab}[1]{#1}
\providecommand{\url}[1]{\texttt{#1}}
\expandafter\ifx\csname urlstyle\endcsname\relax
  \providecommand{\doi}[1]{doi: #1}\else
  \providecommand{\doi}{doi: \begingroup \urlstyle{rm}\Url}\fi

\bibitem[Anderson(1989)]{anderson1989learning}
Anderson, C.~W.
\newblock Learning to control an inverted pendulum using neural networks.
\newblock \emph{IEEE Control Systems Magazine}, 9\penalty0 (3):\penalty0
  31--37, 1989.

\bibitem[Arulkumaran et~al.(2019)Arulkumaran, Cully, and
  Togelius]{arulkumaran2019alphastar}
Arulkumaran, K., Cully, A., and Togelius, J.
\newblock Alphastar: An evolutionary computation perspective.
\newblock \emph{arXiv preprint arXiv:1902.01724}, 2019.

\bibitem[Balduzzi et~al.(2019)Balduzzi, Garnelo, Bachrach, Czarnecki,
  P{\'e}rolat, Jaderberg, and Graepel]{Balduzzi2019OpenendedLI}
Balduzzi, D., Garnelo, M., Bachrach, Y., Czarnecki, W., P{\'e}rolat, J.,
  Jaderberg, M., and Graepel, T.
\newblock Open-ended learning in symmetric zero-sum games.
\newblock \emph{ArXiv}, abs/1901.08106, 2019.

\bibitem[Bansal et~al.(2018)Bansal, Pachocki, Sidor, Sutskever, and
  Mordatch]{selfplay:arxiv18}
Bansal, T., Pachocki, J., Sidor, S., Sutskever, I., and Mordatch, I.
\newblock Emergent complexity via multi-agent competition.
\newblock \emph{arXiv preprint arXiv:1710.03748}, 2018.

\bibitem[Bedau(1992)]{bedau:alife92}
Bedau, M.
\newblock Measurement of evolutionary activity.
\newblock In \emph{Proceedings of the 2nd International Conference on the
  Synthesis and Simulation of Living Systems (ALIFE 2)}, 1992.

\bibitem[Bedau(2008)]{bedau:alife08}
Bedau, M.
\newblock The arrow of complexity hypothesis (abstract).
\newblock In Bullock, S., Noble, J., Watson, R., and Bedau, M. (eds.),
  \emph{Proceedings of the Eleventh International Conference on Artificial Life
  (Alife XI)}, pp.\  750, Cambridge, MA, 2008. MIT Press.
\newblock URL \url{http://www.alifexi.org/papers/ALIFExi-abstracts-0010.pdf}.

\bibitem[Bellemare et~al.(2013)Bellemare, Naddaf, Veness, and
  Bowling]{bellemare2013arcade}
Bellemare, M., Naddaf, Y., Veness, J., and Bowling, M.
\newblock The arcade learning environment: An evaluation platform for general
  agents.
\newblock \emph{Journal of Artificial Intelligence Research}, 47:\penalty0
  253--279, 2013.

\bibitem[Bellemare et~al.(2016)Bellemare, Srinivasan, Ostrovski, Schaul,
  Saxton, and Munos]{bellemare2016unifying}
Bellemare, M., Srinivasan, S., Ostrovski, G., Schaul, T., Saxton, D., and
  Munos, R.
\newblock Unifying count-based exploration and intrinsic motivation.
\newblock In \emph{NIPS}, pp.\  1471--1479, 2016.

\bibitem[Bengio et~al.(2009)Bengio, Louradour, Collobert, and
  Weston]{bengio2009curriculum}
Bengio, Y., Louradour, J., Collobert, R., and Weston, J.
\newblock Curriculum learning.
\newblock In \emph{Proceedings of the 26th annual international conference on
  machine learning}, pp.\  41--48. ACM, 2009.

\bibitem[Brant \& Stanley(2017)Brant and Stanley]{brant:mcc17}
Brant, J.~C. and Stanley, K.~O.
\newblock Minimal criterion coevolution: A new approach to open-ended search.
\newblock In \emph{Proceedings of the 2017 on Genetic and Evolutionary
  Computation Conference (GECCO)}, pp.\  67--74, 2017.

\bibitem[Brockman et~al.(2016)Brockman, Cheung, Pettersson, Schneider,
  Schulman, Tang, and Zaremba]{brockman2016openai}
Brockman, G., Cheung, V., Pettersson, L., Schneider, J., Schulman, J., Tang,
  J., and Zaremba, W.
\newblock Openai gym.
\newblock \emph{arXiv preprint arXiv:1606.01540}, 2016.

\bibitem[Clune(2019)]{clune2019ai}
Clune, J.
\newblock {AI-GA}s: {AI}-generating algorithms, an alternate paradigm for
  producing general artificial intelligence.
\newblock \emph{arXiv preprint arXiv:1905.10985}, 2019.

\bibitem[Cully et~al.(2015)Cully, Clune, Tarapore, and Mouret]{cully:nature15}
Cully, A., Clune, J., Tarapore, D., and Mouret, J.-B.
\newblock Robots that can adapt like animals.
\newblock \emph{Nature}, 521:\penalty0 503--507, 2015.
\newblock \doi{10.1038/nature14422}.

\bibitem[de~Jong \& Pollack(2004)de~Jong and Pollack]{Jong2004IdealEF}
de~Jong, E.~D. and Pollack, J.~B.
\newblock Ideal evaluation from coevolution.
\newblock \emph{Evolutionary Computation}, 12:\penalty0 159--192, 2004.

\bibitem[Deng et~al.(2009)Deng, Dong, Socher, Li, Li, and
  Fei-Fei]{deng2009imagenet}
Deng, J., Dong, W., Socher, R., Li, L.-J., Li, K., and Fei-Fei, L.
\newblock {ImageNet}: A large-scale hierarchical image database.
\newblock In \emph{2009 IEEE conference on computer vision and pattern
  recognition}, pp.\  248--255. IEEE, 2009.

\bibitem[Dhariwal et~al.(2017)Dhariwal, Hesse, Klimov, Nichol, Plappert,
  Radford, Schulman, Sidor, Wu, and Zhokhov]{dhariwal2017openai}
Dhariwal, P., Hesse, C., Klimov, O., Nichol, A., Plappert, M., Radford, A.,
  Schulman, J., Sidor, S., Wu, Y., and Zhokhov, P.
\newblock Openai baselines.
\newblock \emph{GitHub, GitHub repository}, 2017.

\bibitem[Dolson et~al.(2018)Dolson, Vostinar, and Ofria]{dolson2018what}
Dolson, E., Vostinar, A., and Ofria, C.
\newblock What’s holding artificial life back from open-ended evolution?
\newblock \emph{The Winnower}, 04 2018.
\newblock \doi{10.15200/winn.152418.86598}.
\newblock URL \url{https://dx.doi.org/10.15200/winn.152418.86598}.

\bibitem[Duan et~al.(2016)Duan, Schulman, Chen, Bartlett, Sutskever, and
  Abbeel]{duan2016rl}
Duan, Y., Schulman, J., Chen, X., Bartlett, P.~L., Sutskever, I., and Abbeel,
  P.
\newblock Rl $^{} 2$: Fast reinforcement learning via slow reinforcement
  learning.
\newblock \emph{arXiv preprint arXiv:1611.02779}, 2016.

\bibitem[Ecoffet et~al.(2019)Ecoffet, Huizinga, Lehman, Stanley, and
  Clune]{ecoffet2019go}
Ecoffet, A., Huizinga, J., Lehman, J., Stanley, K.~O., and Clune, J.
\newblock Go-explore: a new approach for hard-exploration problems.
\newblock \emph{arXiv preprint arXiv:1901.10995}, 2019.

\bibitem[Ficici \& Pollack(1998)Ficici and Pollack]{ficici:alife98}
Ficici, S. and Pollack, J.
\newblock {Challenges in coevolutionary learning: Arms-race dynamics,
  open-endedness, and mediocre stable states}.
\newblock \emph{Artificial life VI}, pp.\  238, 1998.

\bibitem[Finn et~al.(2017)Finn, Abbeel, and Levine]{finn2017model}
Finn, C., Abbeel, P., and Levine, S.
\newblock Model-agnostic meta-learning for fast adaptation of deep networks.
\newblock In \emph{Proceedings of the 34th International Conference on Machine
  Learning-Volume 70}, pp.\  1126--1135. JMLR. org, 2017.

\bibitem[Florensa et~al.(2017)Florensa, Held, Wulfmeier, Zhang, and
  Abbeel]{pmlr-v78-florensa17a}
Florensa, C., Held, D., Wulfmeier, M., Zhang, M., and Abbeel, P.
\newblock Reverse curriculum generation for reinforcement learning.
\newblock In \emph{Proceedings of the 1st Annual Conference on Robot Learning},
  pp.\  482--495, 2017.

\bibitem[Florensa et~al.(2018)Florensa, Held, Geng, and
  Abbeel]{florensa2018automatic}
Florensa, C., Held, D., Geng, X., and Abbeel, P.
\newblock Automatic goal generation for reinforcement learning agents.
\newblock In \emph{International Conference on Machine Learning}, pp.\
  1514--1523, 2018.

\bibitem[Forestier et~al.(2017)Forestier, Mollard, and
  Oudeyer]{forestier2017intrinsically}
Forestier, S., Mollard, Y., and Oudeyer, P.-Y.
\newblock Intrinsically motivated goal exploration processes with automatic
  curriculum learning.
\newblock \emph{arXiv preprint arXiv:1708.02190}, 2017.

\bibitem[Gomez \& Miikkulainen(1997)Gomez and
  Miikkulainen]{gomez1997incremental}
Gomez, F. and Miikkulainen, R.
\newblock Incremental evolution of complex general behavior.
\newblock \emph{Adaptive Behavior}, 5\penalty0 (3-4):\penalty0 317--342, 1997.

\bibitem[Goodfellow et~al.(2014)Goodfellow, Pouget-Abadie, Mirza, Xu,
  Warde-Farley, Ozair, Courville, and Bengio]{goodfellow2014generative}
Goodfellow, I., Pouget-Abadie, J., Mirza, M., Xu, B., Warde-Farley, D., Ozair,
  S., Courville, A., and Bengio, Y.
\newblock Generative adversarial nets.
\newblock In \emph{Advances in neural information processing systems}, pp.\
  2672--2680, 2014.

\bibitem[Graening et~al.(2010)Graening, Aulig, and Olhofer]{graening:ppsn10}
Graening, L., Aulig, N., and Olhofer, M.
\newblock Towards directed open-ended search by a novelty guided evolution
  strategy.
\newblock In Schaefer, R., Cotta, C., Ko{\l}odziej, J., and Rudolph, G. (eds.),
  \emph{Parallel Problem Solving from Nature -- PPSN XI}, volume 6239 of
  \emph{Lecture Notes in Computer Science}, pp.\  71--80. Springer, 2010.
\newblock ISBN 978-3-642-15870-4.

\bibitem[Heess et~al.(2017)Heess, Sriram, Lemmon, Merel, Wayne, Tassa, Erez,
  Wang, Eslami, Riedmiller, et~al.]{heess2017emergence}
Heess, N., Sriram, S., Lemmon, J., Merel, J., Wayne, G., Tassa, Y., Erez, T.,
  Wang, Z., Eslami, S., Riedmiller, M., et~al.
\newblock Emergence of locomotion behaviours in rich environments.
\newblock \emph{arXiv preprint arXiv:1707.02286}, 2017.

\bibitem[Jaderberg et~al.(2017)Jaderberg, Dalibard, Osindero, Czarnecki,
  Donahue, Razavi, Vinyals, Green, Dunning, Simonyan,
  et~al.]{jaderberg2017population}
Jaderberg, M., Dalibard, V., Osindero, S., Czarnecki, W.~M., Donahue, J.,
  Razavi, A., Vinyals, O., Green, T., Dunning, I., Simonyan, K., et~al.
\newblock Population based training of neural networks.
\newblock \emph{arXiv preprint arXiv:1711.09846}, 2017.

\bibitem[Justesen et~al.(2018)Justesen, Torrado, Bontrager, Khalifa, Togelius,
  and Risi]{justesen2017procedural}
Justesen, N., Torrado, R.~R., Bontrager, P., Khalifa, A., Togelius, J., and
  Risi, S.
\newblock Illuminating generalization in deep reinforcement learning through
  procedural level generation.
\newblock \emph{arXiv preprint arXiv:1806.10729}, 2018.

\bibitem[Karpathy \& Van De~Panne(2012)Karpathy and Van
  De~Panne]{karpathy2012curriculum}
Karpathy, A. and Van De~Panne, M.
\newblock Curriculum learning for motor skills.
\newblock In \emph{Canadian Conference on Artificial Intelligence}, pp.\
  325--330. Springer, 2012.

\bibitem[Kingma \& Ba(2014)Kingma and Ba]{kingma2014adam}
Kingma, D. and Ba, J.
\newblock Adam: A method for stochastic optimization.
\newblock \emph{arXiv preprint arXiv:1412.6980}, 2014.

\bibitem[Klimov(2016)]{OpenAI_bipedal}
Klimov, O.
\newblock Bipedalwalkerhardcore-v2.
\newblock \url{https://gym.openai.com}, 2016.

\bibitem[Langdon(2005)]{langdon:metas05}
Langdon, W.~B.
\newblock Pfeiffer -- {A} distributed open-ended evolutionary system.
\newblock In Edmonds, B., Gilbert, N., Gustafson, S., Hales, D., and Krasnogor,
  N. (eds.), \emph{AISB'05: Proceedings of the Joint Symposium on Socially
  Inspired Computing (METAS 2005)}, pp.\  7--13, 2005.
\newblock URL
  \url{http://www.cs.ucl.ac.uk/staff/W.Langdon/ftp/papers/wbl_metas2005.pdf}.

\bibitem[LeCun et~al.(1998)LeCun, Bottou, Bengio, Haffner,
  et~al.]{lecun1998gradient}
LeCun, Y., Bottou, L., Bengio, Y., Haffner, P., et~al.
\newblock Gradient-based learning applied to document recognition.
\newblock \emph{Proceedings of the IEEE}, 86\penalty0 (11):\penalty0
  2278--2324, 1998.

\bibitem[Lehman \& Stanley(2008)Lehman and Stanley]{lehman:alife08}
Lehman, J. and Stanley, K.~O.
\newblock Exploiting open-endedness to solve problems through the search for
  novelty.
\newblock In Bullock, S., Noble, J., Watson, R., and Bedau, M. (eds.),
  \emph{Proceedings of the Eleventh International Conference on Artificial Life
  (Alife XI)}, Cambridge, MA, 2008. MIT Press.
\newblock URL \url{http://eplex.cs.ucf.edu/papers/lehman_alife08.pdf}.

\bibitem[Lehman \& Stanley(2011{\natexlab{a}})Lehman and
  Stanley]{lehman2011abandoning}
Lehman, J. and Stanley, K.~O.
\newblock Abandoning objectives: Evolution through the search for novelty
  alone.
\newblock \emph{Evolutionary computation}, 19\penalty0 (2):\penalty0 189--223,
  2011{\natexlab{a}}.

\bibitem[Lehman \& Stanley(2011{\natexlab{b}})Lehman and
  Stanley]{lehman:gecco11}
Lehman, J. and Stanley, K.~O.
\newblock Evolving a diversity of virtual creatures through novelty search and
  local competition.
\newblock In \emph{GECCO '11: Proceedings of the 13th annual conference on
  Genetic and evolutionary computation}, pp.\  211--218, 2011{\natexlab{b}}.

\bibitem[Lenski et~al.(2003)Lenski, Ofria, Pennock, and
  Adami]{lenski2003evolutionary}
Lenski, R.~E., Ofria, C., Pennock, R.~T., and Adami, C.
\newblock The evolutionary origin of complex features.
\newblock \emph{Nature}, 423\penalty0 (6936):\penalty0 139, 2003.

\bibitem[Matiisen et~al.(2017)Matiisen, Oliver, Cohen, and
  Schulman]{matiisen2017teacher}
Matiisen, T., Oliver, A., Cohen, T., and Schulman, J.
\newblock Teacher-student curriculum learning.
\newblock \emph{arXiv preprint arXiv:1707.00183}, 2017.

\bibitem[McIntyre et~al.()McIntyre, Kallada, Miguel, and da~Silva]{neat-python}
McIntyre, A., Kallada, M., Miguel, C.~G., and da~Silva, C.~F.
\newblock neat-python.
\newblock \url{https://github.com/CodeReclaimers/neat-python}.

\bibitem[Mouret \& Clune(2015)Mouret and Clune]{mouret:arxiv15}
Mouret, J. and Clune, J.
\newblock Illuminating search spaces by mapping elites.
\newblock \emph{ArXiv e-prints}, abs/1504.04909, 2015.
\newblock URL \url{http://arxiv.org/abs/1504.04909}.

\bibitem[Nguyen et~al.(2016)Nguyen, Yosinski, and
  Clune]{nguyen2016understanding}
Nguyen, A., Yosinski, J., and Clune, J.
\newblock Understanding innovation engines: Automated creativity and improved
  stochastic optimization via deep learning.
\newblock \emph{Evolutionary computation}, 24\penalty0 (3):\penalty0 545--572,
  2016.

\bibitem[OpenAI et~al.(2019)OpenAI, Berner, Brockman, Chan, Cheung, Dębiak,
  Dennison, Farhi, Fischer, Hashme, Hesse, Józefowicz, Gray, Olsson, Pachocki,
  Petrov, de~Oliveira~Pinto, Raiman, Salimans, Schlatter, Schneider, Sidor,
  Sutskever, Tang, Wolski, and Zhang]{openai2019dota}
OpenAI, Berner, C., Brockman, G., Chan, B., Cheung, V., Dębiak, P., Dennison,
  C., Farhi, D., Fischer, Q., Hashme, S., Hesse, C., Józefowicz, R., Gray, S.,
  Olsson, C., Pachocki, J., Petrov, M., de~Oliveira~Pinto, H.~P., Raiman, J.,
  Salimans, T., Schlatter, J., Schneider, J., Sidor, S., Sutskever, I., Tang,
  J., Wolski, F., and Zhang, S.
\newblock Dota 2 with large scale deep reinforcement learning.
\newblock 2019.
\newblock URL \url{https://arxiv.org/abs/1912.06680}.

\bibitem[Oudeyer \& Kaplan(2009)Oudeyer and Kaplan]{oudeyer2009intrinsic}
Oudeyer, P.-Y. and Kaplan, F.
\newblock What is intrinsic motivation? a typology of computational approaches.
\newblock \emph{Frontiers in Neurorobotics}, 1:\penalty0 6, 2009.

\bibitem[Popovici et~al.(2012)Popovici, Bucci, Wiegand, and
  De~Jong]{popovici:hnc12}
Popovici, E., Bucci, A., Wiegand, R.~P., and De~Jong, E.~D.
\newblock \emph{Coevolutionary Principles}, pp.\  987--1033.
\newblock Springer Berlin Heidelberg, Berlin, Heidelberg, 2012.
\newblock ISBN 978-3-540-92910-9.
\newblock \doi{10.1007/978-3-540-92910-9_31}.
\newblock URL \url{http://dx.doi.org/10.1007/978-3-540-92910-9_31}.

\bibitem[Pugh et~al.(2016)Pugh, Soros, and Stanley]{pugh:frontiers16}
Pugh, J.~K., Soros, L.~B., and Stanley, K.~O.
\newblock Quality diversity: A new frontier for evolutionary computation.
\newblock 3\penalty0 (40), 2016.
\newblock ISSN 2296-9144.

\bibitem[Ray(1991)]{ray:alife91}
Ray, T.~S.
\newblock An approach to the synthesis of life.
\newblock In \emph{Artificial Life {II}}, pp.\  371--408. Addison-Wesley, 1991.

\bibitem[Salimans et~al.(2017)Salimans, Ho, Chen, Sidor, and
  Sutskever]{salimans2017evolution}
Salimans, T., Ho, J., Chen, X., Sidor, S., and Sutskever, I.
\newblock Evolution strategies as a scalable alternative to reinforcement
  learning.
\newblock \emph{arXiv preprint arXiv:1703.03864}, 2017.

\bibitem[Samuel(1967)]{samuel1967some}
Samuel, A.~L.
\newblock Some studies in machine learning using the game of checkers.
  ii—recent progress.
\newblock \emph{IBM Journal of research and development}, 11\penalty0
  (6):\penalty0 601--617, 1967.

\bibitem[Schmidhuber(2010)]{schmidhuber2010formal}
Schmidhuber, J.
\newblock Formal theory of creativity, fun, and intrinsic motivation
  (1990--2010).
\newblock \emph{IEEE Transactions on Autonomous Mental Development}, 2\penalty0
  (3):\penalty0 230--247, 2010.

\bibitem[Schmidhuber(2013)]{schmidhuber2013powerplay}
Schmidhuber, J.
\newblock {POWERPLAY}: Training an increasingly general problem solver by
  continually searching for the simplest still unsolvable problem.
\newblock \emph{Frontiers in psychology}, 4:\penalty0 313, 2013.

\bibitem[Schulman et~al.(2017)Schulman, Wolski, Dhariwal, Radford, and
  Klimov]{schulman2017proximal}
Schulman, J., Wolski, F., Dhariwal, P., Radford, A., and Klimov, O.
\newblock Proximal policy optimization algorithms.
\newblock \emph{arXiv preprint arXiv:1707.06347}, 2017.

\bibitem[Shaker et~al.(2016)Shaker, Togelius, and Nelson]{shaker2016procedural}
Shaker, N., Togelius, J., and Nelson, M.~J.
\newblock \emph{Procedural content generation in games}.
\newblock Springer, 2016.

\bibitem[Silver et~al.(2018)Silver, Hubert, Schrittwieser, Antonoglou, Lai,
  Guez, Lanctot, Sifre, Kumaran, Graepel, et~al.]{silver2018general}
Silver, D., Hubert, T., Schrittwieser, J., Antonoglou, I., Lai, M., Guez, A.,
  Lanctot, M., Sifre, L., Kumaran, D., Graepel, T., et~al.
\newblock A general reinforcement learning algorithm that masters chess, shogi,
  and go through self-play.
\newblock \emph{Science}, 362\penalty0 (6419):\penalty0 1140--1144, 2018.

\bibitem[Soros \& Stanley(2014)Soros and Stanley]{soros:alife14}
Soros, L. and Stanley, K.~O.
\newblock Identifying necessary conditions for open-ended evolution through the
  artificial life world of chromaria.
\newblock In \emph{ALIFE 14: The Fourteenth International Conference on the
  Synthesis and Simulation of Living Systems}, pp.\  793--800, 2014.

\bibitem[Soros et~al.(2016)Soros, Cheney, and Stanley]{soros2:alife14}
Soros, L., Cheney, N., and Stanley, K.~O.
\newblock How the strictness of the minimal criterion impacts open-ended
  evolution.
\newblock In \emph{ALIFE 15: The Fifteenth International Conference on the
  Synthesis and Simulation of Living Systems}, pp.\  208--215, 2016.

\bibitem[Standish(2003{\natexlab{a}})]{standish2003open}
Standish, R.~K.
\newblock Open-ended artificial evolution.
\newblock \emph{International Journal of Computational Intelligence and
  Applications}, 3\penalty0 (02):\penalty0 167--175, 2003{\natexlab{a}}.

\bibitem[Standish(2003{\natexlab{b}})]{standish:ijcia03}
Standish, R.~K.
\newblock Open-ended artificial evolution.
\newblock \emph{International Journal of Computational Intelligence and
  Applications}, 3\penalty0 (02):\penalty0 167--175, 2003{\natexlab{b}}.

\bibitem[Stanley(2007)]{stanley2007compositional}
Stanley, K.~O.
\newblock Compositional pattern producing networks: A novel abstraction of
  development.
\newblock \emph{Genetic programming and evolvable machines}, 8\penalty0
  (2):\penalty0 131--162, 2007.

\bibitem[Stanley \& Lehman(2015)Stanley and Lehman]{stanley:book15}
Stanley, K.~O. and Lehman, J.
\newblock Why greatness cannot be planned.
\newblock 2015.

\bibitem[Stanley \& Miikkulainen(2002)Stanley and Miikkulainen]{stanley:ec02}
Stanley, K.~O. and Miikkulainen, R.
\newblock Evolving neural networks through augmenting topologies.
\newblock \emph{Evolutionary Computation}, 10\penalty0 (2):\penalty0 99--127,
  2002.
\newblock URL \url{http://nn.cs.utexas.edu/?stanley:ec02}.

\bibitem[Stanley et~al.(2017)Stanley, Lehman, and Soros]{stanley2017open}
Stanley, K.~O., Lehman, J., and Soros, L.
\newblock Open-endedness: The last grand challenge you’ve never heard of.
\newblock \emph{O'Reilly Online}, 2017.

\bibitem[Stanley et~al.(2019)Stanley, Clune, Lehman, and
  Miikkulainen]{stanley2019nature}
Stanley, K.~O., Clune, J., Lehman, J., and Miikkulainen, R.
\newblock Designing neural networks through neuroevolution.
\newblock \emph{Nature Machine Intelligence}, 1\penalty0 (1):\penalty0 24--35,
  2019.

\bibitem[Taylor et~al.(2016)Taylor, Bedau, Channon, Ackley, Banzhaf, Beslon,
  Dolson, Froese, Hickinbotham, Ikegami, McMullin, Packard, Rasmussen, Virgo,
  Agmon, Clark, McGregor, Ofria, Ropella, Spector, Stanley, Stanton, Timperley,
  Vostinar, and Wiser]{taylor:alife16}
Taylor, T., Bedau, M., Channon, A., Ackley, D., Banzhaf, W., Beslon, G.,
  Dolson, E., Froese, T., Hickinbotham, S., Ikegami, T., McMullin, B., Packard,
  N., Rasmussen, S., Virgo, N., Agmon, E., Clark, E., McGregor, S., Ofria, C.,
  Ropella, G., Spector, L., Stanley, K.~O., Stanton, A., Timperley, C.,
  Vostinar, A., and Wiser, M.
\newblock Open-ended evolution: Perspectives from the {OEE} workshop in {York}.
\newblock \emph{Artificial life}, 22\penalty0 (3):\penalty0 408–423, 2016.

\bibitem[Togelius et~al.(2011)Togelius, Yannakakis, Stanley, and
  Browne]{togelius:ieeetciaig11}
Togelius, J., Yannakakis, G.~N., Stanley, K.~O., and Browne, C.
\newblock Search-based procedural content generation: A taxonomy and survey.
\newblock \emph{{IEEE} Transactions on Computational Intelligence and AI in
  Games}, 3\penalty0 (3):\penalty0 172--186, 2011.

\bibitem[Vinyals et~al.(2019)Vinyals, Babuschkin, Czarnecki, Mathieu, Dudzik,
  Chung, Choi, Powell, Ewalds, Georgiev, et~al.]{vinyals2019grandmaster}
Vinyals, O., Babuschkin, I., Czarnecki, W.~M., Mathieu, M., Dudzik, A., Chung,
  J., Choi, D.~H., Powell, R., Ewalds, T., Georgiev, P., et~al.
\newblock Grandmaster level in {StarCraft II} using multi-agent reinforcement
  learning.
\newblock \emph{Nature}, 575\penalty0 (7782):\penalty0 350--354, 2019.

\bibitem[Wang et~al.(2016)Wang, Kurth{-}Nelson, Tirumala, Soyer, Leibo, Munos,
  Blundell, Kumaran, and Botvinick]{wang2016lRL}
Wang, J.~X., Kurth{-}Nelson, Z., Tirumala, D., Soyer, H., Leibo, J.~Z., Munos,
  R., Blundell, C., Kumaran, D., and Botvinick, M.
\newblock Learning to reinforcement learn.
\newblock \emph{CoRR}, abs/1611.05763, 2016.
\newblock URL \url{http://arxiv.org/abs/1611.05763}.

\bibitem[Wang et~al.(2019{\natexlab{a}})Wang, Lehman, Clune, and
  Stanley]{wang2019paired}
Wang, R., Lehman, J., Clune, J., and Stanley, K.~O.
\newblock {Paired Open-Ended Trailblazer} ({POET}): Endlessly generating
  increasingly complex and diverse learning environments and their solutions.
\newblock \emph{arXiv preprint arXiv:1901.01753}, 2019{\natexlab{a}}.

\bibitem[Wang et~al.(2019{\natexlab{b}})Wang, Lehman, Clune, and
  Stanley]{wang2019poetGECCO}
Wang, R., Lehman, J., Clune, J., and Stanley, K.~O.
\newblock {POET}: open-ended coevolution of environments and their optimized
  solutions.
\newblock In \emph{Proceedings of the Genetic and Evolutionary Computation
  Conference}, pp.\  142--151. ACM, 2019{\natexlab{b}}.

\bibitem[Wiegand et~al.(2001)Wiegand, Liles, and Jong]{wiegand:gecco01}
Wiegand, R.~P., Liles, W.~C., and Jong, K. A.~D.
\newblock An empirical analysis of collaboration methods in cooperative
  coevolutionary algorithms.
\newblock In \emph{Proceedings of the 3rd Annual Conference on Genetic and
  Evolutionary Computation}, GECCO'01, pp.\  1235--1242, San Francisco, CA,
  USA, 2001. Morgan Kaufmann Publishers Inc.
\newblock ISBN 1-55860-774-9.
\newblock URL \url{http://dl.acm.org/citation.cfm?id=2955239.2955458}.

\bibitem[Wierstra et~al.(2014)Wierstra, Schaul, Glasmachers, Sun, Peters, and
  Schmidhuber]{wierstra2014natural}
Wierstra, D., Schaul, T., Glasmachers, T., Sun, Y., Peters, J., and
  Schmidhuber, J.
\newblock Natural evolution strategies.
\newblock \emph{The Journal of Machine Learning Research}, 15\penalty0
  (1):\penalty0 949--980, 2014.

\bibitem[Yosinski et~al.(2014)Yosinski, Clune, Bengio, and
  Lipson]{yosinski2014transferable}
Yosinski, J., Clune, J., Bengio, Y., and Lipson, H.
\newblock How transferable are features in deep neural networks?
\newblock In \emph{Advances in neural information processing systems}, pp.\
  3320--3328, 2014.

\bibitem[Zhi et~al.(2020)Zhi, Wang, Clune, and Stanley]{zhi2020fiber}
Zhi, J., Wang, R., Clune, J., and Stanley, K.~O.
\newblock Fiber: A platform for efficient development and distributed training
  for reinforcement learning and population-based methods.
\newblock \emph{arXiv preprint arXiv:2003.11164}, 2020.

\end{thebibliography}
\bibliographystyle{icml2020}

\clearpage
\appendix
\section{Appendix}

\subsection{Algorithmic Description of the Improved Transfer Strategy in Section \ref{innovations}}
\label{appendix:improved_transfer_algo}
\begin{algorithm}[htp]
\caption{Improved Transfer}
\label{alg:eff_transfer}
\begin{algorithmic}[1]
   \STATE {\bfseries Input:} candidate agents denoted by their policy parameter vectors $\theta^{1}$, $\theta^{2}$, \ldots, $\theta^{M}$, target environment $E$ with a score function $Score(\cdot)$, and \texttt{threshold} (i.e.\ max of 5 most recent scores of the incumbent agent)
   \STATE {\bfseries Initialize:} Set list \texttt{P\_Candidates} empty 
   \FOR{$m=1$ {\bfseries to} $M$}
   \STATE Compute direct transfer $D$
   \IF{$Score$($D$) $>$ \texttt{threshold}} 
      \STATE Compute fine-tuning transfer $P$
      \IF{$Score$($P$) $>$ \texttt{threshold}} 
        \STATE add  $P$ to \texttt{P\_Candidates}
      \ENDIF
   \ENDIF
   \ENDFOR
   \STATE {\bfseries Return:} $\argmax_{\theta \in \texttt{P\_Candidates}} Score(\theta) $

\end{algorithmic}
\end{algorithm}

\subsection{Training CPPNs with NEAT}
\label{appendix:cppn-details}

CPPNs are trained with the NEAT algorithm \cite{stanley:ec02}, a neuroevolution \cite{stanley2019nature} algorithm that learns both the architecture (i.e.\ the topology) and the weights of CPPNs. Specifically, in this work, the NEAT-Python library \cite{neat-python} is used to initialize
and evolve the CPPNs that encode environments. The setup and choices of hyperparameter are listed in Table \ref{tab:cppn_hyperparams}. Because POET has its own diversity preservation mechanism, NEAT is run in POET without its conventional speciation mechanism. In this work, crossover between CPPNs is not performed.

\begin{table*}[!ht]
\centering
\begin{tabular}{lr}
\toprule
Hyperparameter & Setting \\
\midrule
initial connection & full \\
activation default & random \\
activation options & identity sin sigmoid square tanh \\
aggregation default & sum \\
bias init stdev & 0.1 \\
bias init type & gaussian \\
bias max value & 10.0 \\
bias min value & -10.0 \\
bias mutate power & 0.1 \\
bias mutate rate & 0.75 \\
compatibility disjoint coefficient & 1.0 \\
compatibility weight coefficient & 0.5 \\
enabled default & True \\
feed forward & True \\
node add prob & 0.1 \\
node delete prob & 0.075 \\
num inputs & 1 \\
num outputs & 1 \\
response init mean & 1.0 \\
response init type & gaussian \\
response max value & 10.0 \\
response min value & -10.0 \\
response mutate power & 0.2 \\
single structural mutation & True \\
structural mutation surer & default \\
weight init stdev & 0.25 \\
weight init type & gaussian \\
weight max value & 10.0 \\
weight min value & -10.0 \\
weight mutate power & 0.1 \\
weight mutate rate & 0.75 \\
\bottomrule
\end{tabular}
\caption{The setup and hyperparameter values for instantiating and evolving CPPNs.}
\label{tab:cppn_hyperparams}
\end{table*}

\subsection{Additional Information about the Domain and Experiment Setup}
\label{appedix:exp_setup}

\subsubsection{Additional Details about the Domain}
\label{appenidx:additional_domain_details}
The agent, illustrated in Figure \ref{fig:cppn}, has four degrees of freedom (i.e.\ the dimensions of its action space) as the hips and knees of each leg are controlled by two motors. It has a total of 24 inputs: readings from 10 LIDAR rangefinders and 14 positional and movement variables from the agent's body parts \cite{OpenAI_bipedal}.

Reward is accumulated at each step when the agent attempts to move from the left end to the right end of an environment. If the agent falls at any step, the reward for that step is $-100$. As long as it does not fall, the step-wise reward is $130 \times \Delta x-5\times\Delta \texttt{Hull\_Angle} - 3.5e-4\times \texttt{Applied\_Torque}$, which encourages the agent to move forward while keeping the hull straight and minimizing motor torque applied at joints. 

An episode terminates when 2,000 time steps (frames) have elapsed, when the agent's head touches any obstacle or ground, or when it arrives at the finish line (the right end of the obstacle course). 

Following \citet{wang2019paired,wang2019poetGECCO}, an environment is considered \emph{solved} when the agent both reaches the finish line and obtains a score of 230 or above. The controller (with 24 inputs and 4 outputs all bounded between \text{-}1 and 1) is a fully-connected neural network with with 2 hidden layers of 40 units each, with $tanh$ activation  functions. 

\subsubsection{POET Experiment Setup}

The hyperparameters for ES used in POET,  and later in controls when relevant, are listed in Table \ref{tab:poet_hyperparams}. POET attempts to generate new environments every $150$ iterations ($M$ in step (1) in Section \ref{overview_poet_framework}), and conducts transfer evaluation experiments every $25$ iterations ($N$ in step (3) in Section \ref{overview_poet_framework}). When any environment-agent pair accepts a transfer or when a child environment-agent pair is first created, the state of its Adam optimizer, the learning rate, standard deviation for noise are reset to their initial values, respectively. 

\begin{table*}[htp]
\centering
\begin{tabular}{lr}
\toprule
Hyperparameter & Setting \\
\midrule
number of sample points for each ES step & 512 \\
weight update method & Adam \cite{kingma2014adam} \\
initial learning rate & 0.01 \\
lower bound of learning rate  & 0.001 \\
decay factor of learning rate  per ES step &  0.9999 \\
initial noise standard deviation for ES & 0.1 \\
lower bound of noise standard deviation  & 0.01 \\
decay factor of noise standard deviation  per ES step &  0.999 \\

\bottomrule
\end{tabular}
\caption{Hyperparameters for ES in POET experiments and controls.}
\label{tab:poet_hyperparams}
\end{table*}

\subsection{Additional Details and Results on Evaluation of Algorithmic Innovations}
\label{appendix:exp_1_2}

\subsubsection{Performance Comparison between PATA-EC and the Oracle EC}
\label{appendix:figure_4}

\begin{figure*}[!ht]
  \centering
  \includegraphics[width=.6\linewidth]{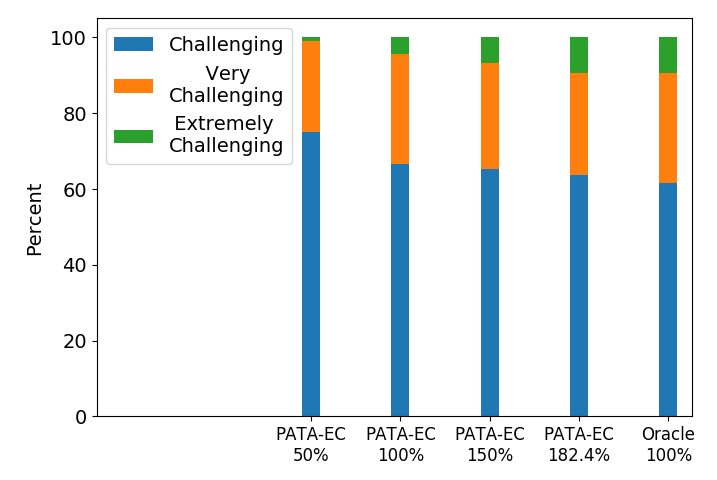}
\caption{
\textbf{Performance comparison between PATA-EC and the oracle EC when applied to the environment in the original POET (with the hand-crafted simple EE).} The domain-general PATA-EC can match the ability of the hand-designed, domain-specific Oracle EC to generate diverse environments of different challenge levels, although it requires more computation to do so. The number below each treatment reports the fraction of computation that treatment received relative to that with the Oracle EC (100\%). As compute increases, POET with the domain-general PATA-EC can generate increasingly diverse, challenging environments, eventually matching the hand-designed, domain-specific Oracle EC in that regard. Definitions of ``\emph{challenging}'', ``\emph{very challenging}'', and ``\emph{extremely challenging}'' environments are the same as those in the original POET \cite{wang2019paired, wang2019poetGECCO} and are also explained in Section \ref{experiments}).
}
  \label{fig:gebc_vs_oracle}
\end{figure*}

Recall that PATA-EC is general and can be applied to nearly any environment and with any encoding.
The idea in this experiment is to apply the new PATA-EC to the environment in the original POET (which still uses the original, hand-crafted EE), and then to compare the result to POET's performance with the original hand-designed EC on the same hand-crafted EE. The question is whether the more generic PATA-EC can perform reasonably close to an EC explicitly hand-crafted for this domain. With this setup, holding everything the same as in the original POET paper except for the EC, we find that the general PATA-EC can indeed produce the same diversity and levels of challenge environments as the original hand-designed EC, although it is less efficient at doing so. It requires $82.4 \pm 7.31 \%$ more computation, measured in ES steps, to produce the same level of complexity (Figure \ref{fig:gebc_vs_oracle}).   

\subsubsection{Performance Comparison between Different Transfer Strategies}
\label{appendix:figure_5}

\begin{figure*}[!ht]
  \centering
  \includegraphics[width=.6\linewidth]{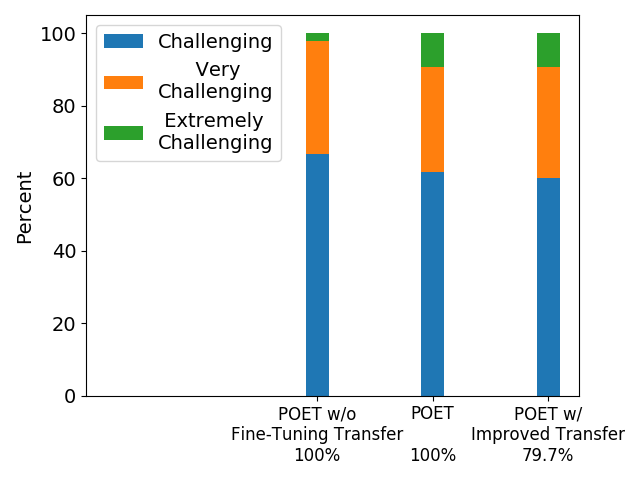}
\caption{\textbf{Percentage of environments with different challenge levels created and solved by POET with different transfer strategies.}  With the improved transfer strategy, POET is able to create and solve environments with the same diversity and challenge levels as the original POET with less computation. In contrast, removing all fine-tuning transfer performs poorly. The number below each treatment reports the fraction of computation that treatment received relative to that for original POET ($100\%$).
}
  \label{fig:compare-transfers}
\end{figure*}

As shown in \citet{wang2019paired, wang2019poetGECCO}, periodic transfer attempts are essential to obtaining solutions in extremely challenging environments, despite being computationally expensive (Section \ref{innovations}). Here the different transfer strategies are compared in the same setup as in original POET \cite{wang2019paired, wang2019poetGECCO}.
POET with the improved transfer strategy creates (and solves) the same fraction of extremely challenging environments and achieves a similar diversity and challeng levels as the original POET, but with only $79.7\pm1.67\%$ of the computation (measured in number of ES steps) of the original POET (Figure \ref{fig:compare-transfers}). Furthermore, the chance of an existing agent paired with an environment being replaced by a transferred agent from another environment dropped from $50.44\pm3.39\%$ with the original POET (this high number suggests many false positives) to $22.31\pm2.42\%$. For comparison, the corresponding number from Innovation Engines \cite{nguyen2016understanding} (which exhibited healthy goal-switching and optimization dynamics) was a similar $17.9\%$. 
A simple alternative, which is removing all fine tuning (i.e.\ not running ES at all as part of a transfer attempt), performs poorly (Figure \ref{fig:compare-transfers}). These results justify that some fine tuning is indeed necessary for finding promising stepping stones. They also suggest that there are efficiency gains when only paying the computational cost of such fine tuning once the direct transfer test is satisfied. 

\subsection{Sample Environments}
\label{appendix:app_landscape_samples}
The 12 sample environments illustrated in Figure \ref{fig:landscape_12}   that were created and solved by Enhanced POET with the CPPN-based EE in a single run are selected based on the following procedure: Let set $\mathcal{A}$ contain all the environments that POET created and solved in a run, and let $\mathcal{S}$ be initialized as a single-element set that contains only the perfectly flat environment, i.e., the very first environment that any POET run starts with. At each iteration of this procedure, we add to $\mathcal{S}$ environment $E$ that satisfies 
$\argmax_{E \in \mathcal{A}, E \notin \mathcal{S}} \min_{e \in S} \mathcal{D}(E, e)$, where $\mathcal{D}(\cdot,\cdot)$ measures the distance between any given two environment. Here we adopt the same distance measure based on PATA-EC as that used in novelty calculation as stated in Section \ref{innovations}. We repeat the iterative process until number of environments other than the perfect flat environment in $\mathcal{S}$ reaches a preset value. 

The intuition behind this selection process is that we continually add the environment that is furthest away from those in $\mathcal{S}$ using the distance measure proposed in this work. It is evident that those environments in Figure \ref{fig:landscape_12} exhibit a broad variety of obstacles that are vastly diverse not only in overall shapes and heights but also in fine details and local variations. This collection is a validation of the diversity-promoting nature of POET, and also a validation that the proposed distance metric based on PATA-EC does help to capture how different the environments are from each other.

\begin{figure*}
  \centering
  \includegraphics[width=1.\linewidth]{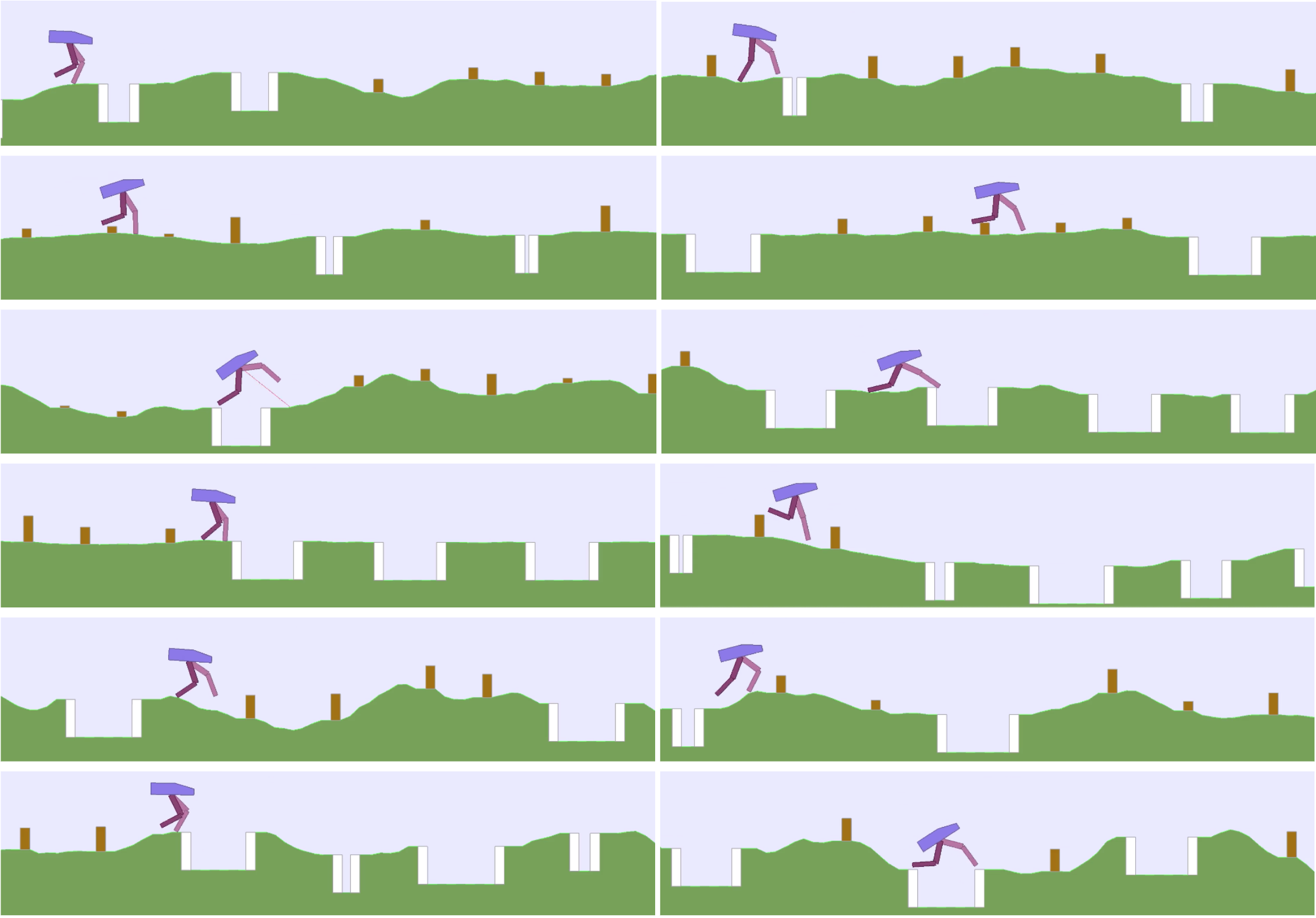}
\caption{\textbf{Sample environments in original POET.} The simple, hand-designed EE can only support a finite set of types of obstacles, i.e., rough surfaces, stumps with fixed width and variable heights, and gaps with variable widths. This search space sustains some, but limited, diversity. 
Each column shows six sample environments from one run, where the upper, middle, and bottom two rows illustrate two ``challenging'', ``very challenging'', and ``extremely challenging'' environments, respectively, according to the challenge levels defined in Section \ref{experiments}.
}
  \label{fig:original_landscape_12}
\end{figure*}

For comparison with the CPPN-encoded environments illustrated in Figure \ref{fig:landscape_12}, Figure \ref{fig:original_landscape_12} illustrates 12 sample environments that were created and solved in original POET with the simple, hand-designed encoding that only supports surface roughness and two regularly-shaped obstacles, i.e., stumps and gaps. Each column illustrates six sample environments from one run of original POET, where the upper, middle, bottom two rows illustrate the environments categorized as ``challenging'', ``very challenging'', and ``extremely challenging'' environments, respectively as defined in Section \ref{experiments}.

\subsection{Phylogenetic Tree}
\label{appendix:phylogenetic_tree}

One way to visualize the diversity POET produces is by viewing a phylogenetic tree (i.e.\ a family tree) of the environments it has created at any given point. Natural phylogenetic trees have numerous, deep, nested branches. For example, nature has many phlya (e.g. mammals, plants, fungi, bacteria, etc.), each of which has many different, branches within it that are long (in that they have persisted over long periods of evolutionary time). Historic attempts at creating open-ended explosions of complexity
in computer simulations of evolving systems in the fields of artificial life and computational evolutionary biology rarely, if ever, produce such phylogenetic trees~\cite{lenski2003evolutionary}. Instead, usually one type of agent becomes more fit than everything else and replaces all the other types of agents, eliminating diversity (i.e. pruning all branches of the tree save one). These trees thus have one long trunk and a few shallow branches at the end that capture the not-yet-wiped out diversity in the current population. 

Phylogenetic trees produced by POET, in sharp contrast, more resemble those from nature. Figure \ref{fig:phy_tree} shows the phylogenetic tree of the first 100 environments of a POET run. Each node corresponds a unique environment that POET created with an inserted picture illustrating its landscape. An edge connects an environment (on the upper end) to its child environment (on the lower end). The shape of nodes distinguishes whether environments are still in the active population (circular) or already in the archive (square) at the iteration when the 100th environment is added to the population, while the color of the border of each node suggests its time of creation: darker color means being created later in the process. Note the hierarchical organization, with major families, families within those, etc. The signature of open-ended algorithms is there: complex phylogenetic trees, meaning those that have multiple, deep, hierarchically nested branches. Of course, this tree is much smaller than those in nature, but that could be a function of (1) limited computational resources, and (2) that the environment search space in these first experiments with Enhanced POET are limited to obstacle courses only. Both are subjects in the discussion in Section \ref{conclusion}.

The red arrows indicate when successful transfers happened (i.e.\ existing paired agents being replaced by transferred agents after fine-tuning) during one of transfer iterations (after the 100th environment is added, but before the 101th environment is added). Although most successful transfers happen between neighboring (more similar) environments, some agents manage to transfer to environments that are far from their paired environments (long red arrows in Figure \ref{fig:phy_tree}).

\subsection{Illustration of the Generalization Ability of Agents}
\label{appendix:pata_matrix}

\begin{figure*}
  \centering
  \includegraphics[width=.75\linewidth]{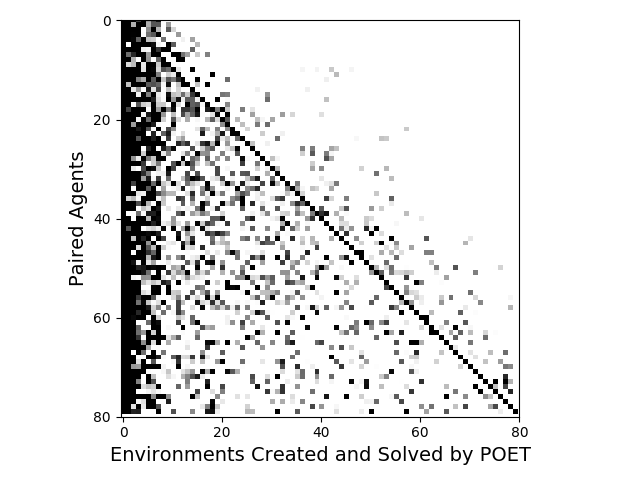}
\caption{\textbf{Illustration of the vectors of scores of how all agents perform across all the first 80 environments that a POET run creates and solves.} Environments are ordered from left to right by the time of creation, while rows corresponds to their respective paired agents. Each square in the plot indicates the normalized score that the agent in that row performs in the environment of its respective column. The normalized score is between 0.0 and 1.0, and color-coded linearly as a grayscale between white and black. Normalization of raw scores is described in Appendix  \ref{appendix:pata_matrix}. }
  \label{fig:heatmap}
\end{figure*}

 
Figure \ref{fig:heatmap} illustrates the vectors of scores for how all agents perform across all the first 80 environments that a POET run creates and solves. More specifically, the $i$th column from left illustrates the vector of scores of all agents evaluated in the $i$th environment numbered in the order of being created, while the $i$th row from the top indicate the performance of the agent paired with the $i$th environment when tested across all the environments, respectively. 

For the purpose of illustration, the raw score is normalized by 230, the minimum score POET has to achieve to solve the respective target environment \cite{wang2019paired,wang2019poetGECCO}, and clipped between 0.0 and 1.0. That way, a normalized score of 1.0 is equivalent to an agent \emph{solving} the environment, a normalized score of 0.0 means the agent achieved a zero or negative score in the environment, and a normalized score between 0.0 and 1.0 indicates that the agent makes some progress, but ultimately fails to solve the environment. The color intensity of matrix entries linearly scales with the normalized scores with white for 0.0 and black for 1.0.

In Figure \ref{fig:heatmap}, all diagonal entries are 1.0 because all environments shown are solved by their paired agents, while any given row reflects the ability of the corresponding agent to generalize across different environment. The upper-right triangular portion indicates how agents perform in environments created later than their paired environments. As environments created later are often more challenging, it makes sense that agents would perform poorly in the environments created much later than their paired ones (indicated by areas towards upper right). As a result, that area is mostly white with some light gray squares near the diagonal. It is also interesting to see that, based on the lower-left triangular portion, agents have limited success in environments that are created earlier than their paired environment as well. This phenomenon indicates that there are no universal ``generalists'' created by POET that are capable of solving all or most of the environments. Instead, over time POET creates ``specialists'' that are mostly specialized to their paired environments.

\subsection{Additional Details and Results about Direct Optimization Control and the Ground-Interpolation Curriculum Control}

\label{appendix:ground-integ-curriculum}

\subsubsection{PPO Experiment Setup}
\label{appendix:PPO_setup}
We adopt the PPO2 implementation from OpenAI Baselines \cite{dhariwal2017openai}. The controller consists of a policy network and a value network. The policy network has the same architecture and activation functions as those used in POET (see \ref{appenidx:additional_domain_details}).  The value network shares the input and the hidden layers with the policy network and it has a separate fully-connected layer that connects to the value output. Hyperparameters listed in Table \ref{tab:ppo_hyperparams} are chosen based on a grid search that yields the highest average scores across three environments randomly sampled from all target environments. We then hold this set of hyperparameters for all the PPO runs for all the target environments. Note that, as illustrated in Figure \ref{fig:percentage_of_solved}, PPO with these hyperparameters has effectively solved early-stage and some middle-stage target environments either by direct optimization or through the ground integration curriculum.   

\begin{table*}[htp]
\centering
\begin{tabular}{lr}
\toprule
Hyperparameter & Setting \\
\midrule
batch size & 65536 \\
number of training minibatches per update & 4 \\
number of training epochs per update & 4 \\
$\lambda$ & 0.95 \\
$\gamma$ & 0.99 \\
value function loss coefficient & 0.5 \\
gradient norm clipping coefficient & 0.5 \\
learning rate & 0.0003 \\
learning rate schedule &  Anneal linearly to 0 \\
\bottomrule
\end{tabular}
\caption{Hyperparameters for PPO experiments.}
\label{tab:ppo_hyperparams}
\end{table*}

\subsubsection{Equivalent Computational Budget}
\label{appendix:equal_budget}

For both direct optimization control and the ground interpolation curriculum control, each run is given the same computational budget as POET spent to solve the target environment, measured in total number of time steps in simulation. It also includes all the simulation rollouts taken in order for POET to solve all the ancestor environments on the direct line leading to the target environment and  all the computations related to transfer attempts into those environments.  

\subsubsection{Other Details and Results}
\label{appendix:additional_control}

As described in Section \ref{experiments}, the 15 target environments were sampled from the three different stages of POET runs. 
That is, for each target environment, we attempted 5 independent runs from different random seeds using direct optimization with ES, direct optimization with PPO, ground-integration curriculum control with ES, and ground-integration curriculum with PPO, respectively (for a total of 300 runs for all target enviroments). 

Figure \ref{fig:continuous_direct_control} reports the normalized scores (following the same normalization method in Appendix \ref{appendix:pata_matrix}) obtained by direct optimization for target environments at different stages. As with environments discovered by original POET, direct optimization can only solve target environments selected at the earlier stages of a POET run (when the produced environments are often less challenging), but neither ES nor PPO could solve more challenging target environments selected at later stages of POET runs. The normalized scores of direct optimization on middle and late stage target environments are significantly lower than 1.0 ($p < 0.01$; Wilcoxon signed rank test).  

The ground interpolation curriculum control follows a setup similar to that of the direct-path curriculum-building control in the original POET \cite{wang2019paired, wang2019poetGECCO}. For each run, the agent starts in a perfectly flat environment. When in one environment the agent achieves a score above the reproduction eligibility threshold of POET (i.e.\ the condition for when an environment-agent pair is eligible to reproduce in POET),  it moves to the next environment (whose scaling factor is increased by 0.02 from the current one). The run stops when the agent reaches and solves the target environment, or when the computational budget (Appendix \ref{appendix:equal_budget}) is used up.

\begin{figure*}[!ht]
  \centering
    \includegraphics[width=.6\linewidth]{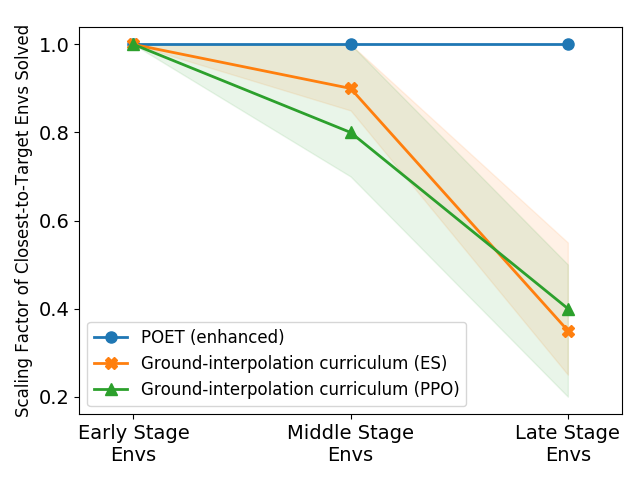}
    \caption{\textbf{Scaling factors of last solved environments on the ground-interpolation curriculum towards target environments.} Symbols: median. Shaded regions: $95\%$ bootstrapped confidence intervals.}
    \label{fig:continuous_curr_control}
\end{figure*}

Figure \ref{fig:continuous_curr_control} illustrates the scaling factor of the last-solved environment by the ground-interpolation curriculum that is closest to the target environment along the path. Statistical tests demonstrate that the ground-interpolation curriculum controls significantly underperform POET in solving late-stage target environment ($p < 0.01$; Wilcoxon signed rank test).

\begin{figure*}[!ht]
  \centering
  \includegraphics[width=.6\linewidth]{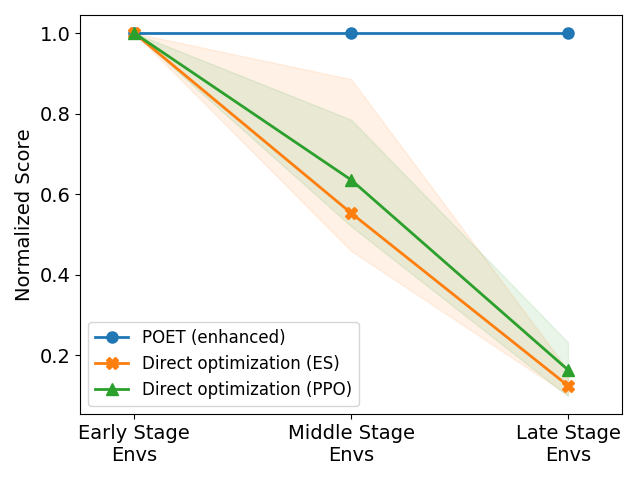}
   \caption{\textbf{Normalized scores of direct optimization on target environments.} Symbols: median. Shaded regions: $95\%$ bootstrapped confidence interval.}
   \label{fig:continuous_direct_control}
\end{figure*}

\end{document}